\documentclass[10pt,twocolumn,letterpaper]{article}

\PassOptionsToPackage{table}{xcolor}

\usepackage[pagenumbers]{cvpr}   

\usepackage{multirow}

\definecolor{cvprblue}{rgb}{0.21,0.49,0.74}
\usepackage[pagebackref,breaklinks,colorlinks,allcolors=cvprblue]{hyperref}

\def\paperID{*****} 
\def\confName{CVPR}
\def\confYear{2026}

\title{MOSAIC: Adaptive Inter-layer Composition for Efficient\\Heterogeneous Vision-Language Models}

\author{Yuncheng Yang\thanks{Equal contribution.}\;\;\; Feiyang Ye\footnotemark[1]\;\;\; Shixian Luo\thanks{Corresponding author.}\;\;\; Yinna Zhu\;\;\;  Lianlei Shan\;\;\; \\ Wangcai Zhao\;\;\; Kuo Zhang \;\;\; Yan Chen \;\;\; Yong Wu\footnotemark[2] \;\;\;  Yan Xie \;\;\;
\vspace{2mm}\\
LiAuto Inc.\\
}

\begin{document}
\maketitle
\begin{abstract}
{\small
Vision-Language Models (VLMs) have achieved significant success by employing homogeneous Transformer architectures to process multimedia information, specifically visual and textual data. Recent studies on Large Multimodal Models (LMMs) indicate that heterogeneous structures interleaving efficient mechanisms, such as linear attention, have demonstrated improvements in both performance and inference latency compared to homogeneous designs.
However, these efforts rely on handcrafted designs with static mixing patterns, which are inherently sub-optimal and difficult to adapt to specific hardware deployment targets. To bridge this gap, we propose \textbf{M}ulti-\textbf{O}bjective \textbf{S}earch for \textbf{A}daptive \textbf{I}nter-layer \textbf{C}omposition (MOSAIC), a hardware-aware search method that automatically transforms homogeneous models into optimized heterogeneous architectures. MOSAIC integrates diverse efficiency mechanisms, including linear, sparse, and low-rank operators, into a unified search space. By formulating the selection process as a multi-objective Mixed Integer Programming (MIP) problem, our method identifies optimal configurations that maximize downstream performance under strict hardware latency constraints. To mitigate the performance degradation arising from structural transitions, we introduce a two-stage parameter recovery process. We first perform global off-policy distillation to stabilize the model's internal representations, followed by a dual-teacher on-policy distillation strategy that leverages a 235B oracle teacher for knowledge expansion while utilizing the original 4B teacher to maintain distributional stability.
We validate the effectiveness of MOSAIC through MOSAIC-4B, a heterogeneous model derived from Qwen3-VL-4B-Instruct. Experimental results demonstrate that MOSAIC-4B matches the performance of the Qwen3-VL-4B-Instruct baseline across multiple benchmarks while requiring less than 2\% of the training cost of the original model. Furthermore, MOSAIC-4B substantially improves inference efficiency, achieving a $1.76\times$ prefilling speedup and $2.54\times$ decoding acceleration. Our MOSAIC-4B model is publicly available at \url{https://huggingface.co/LiAuto-DSR/MOSAIC-4B}}.
\end{abstract}

\section{Introduction}
\label{sec:intro}
Vision-Language Models (VLMs) \cite{wu2025qwen, zhang2024vision, guo2025seed1, liu2023visual, radford2021learning} play a pivotal role in modern real-world multimodal systems, ranging from autonomous driving \cite{yuan2025autodrive, zhou2025autovla, li2025recogdrive} to embodied AI \cite{kim2025fine, xiao2025ava, black2024pi_0, zhang2025dreamvla}. Standard VLM designs predominantly rely on homogeneous architectures, which feature a fixed sequential interleaving of dense self-attention and multi-layer perceptron (MLP) layers. However, the scalability of these architectures is significantly challenged by the quadratic time complexity of standard attention mechanisms \cite{vaswani2017attention}. This complexity results in prohibitive computational overhead, particularly in long-context scenarios \cite{gu2024mamba, lieber2024jamba} or under specific hardware deployment constraints.

\begin{figure}[t]
\centering
\vspace{0.1cm}
\includegraphics[width=\columnwidth]{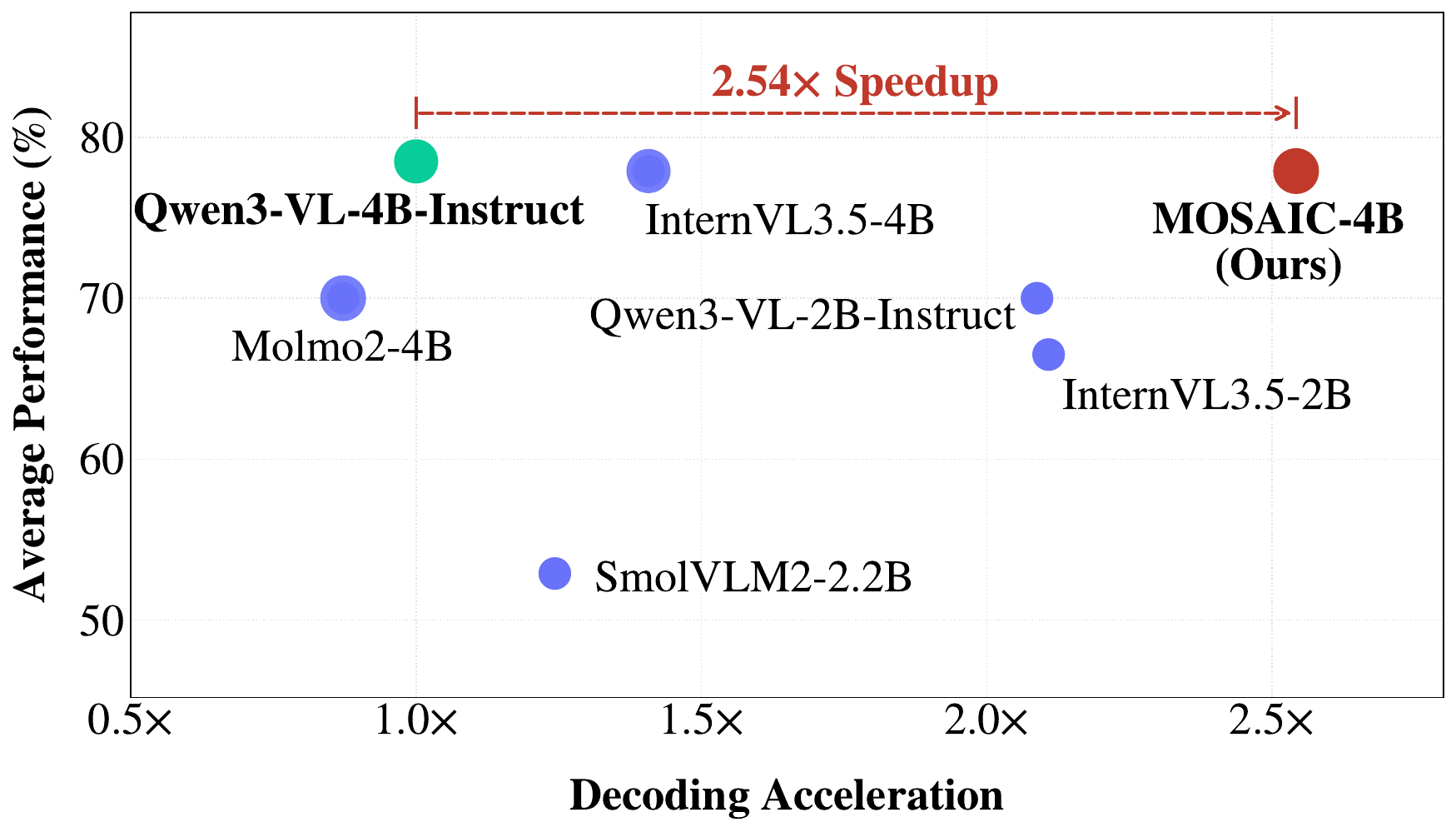}
\caption{Average performance on image understanding benchmarks vs.\ decoding acceleration, measured by time per output token (TPOT). MOSAIC-4B matches the performance of the teacher model (Qwen3-VL-4B-Instruct) while achieving a $2.54\times$ TPOT speedup.
}
\vspace{-0.8in}
\label{fig:head_fig}
\end{figure}

Recent studies on LMMs indicate that heterogeneous structures, which interleave efficient mechanisms such as linear attention and sparse attention, demonstrate significant advantages in both quality and efficiency over purely homogeneous designs \cite{team2025kimi, tao2025infinitevl, lenz2025jamba, waleffe2024empirical, li2025minimax, li2025matvlm}. For instance, the Kimi Linear architecture \cite{team2025kimi} utilizes Kimi Delta Attention (KDA) to outperform traditional full-attention models in long-context scenarios while significantly reducing the memory overhead of the KV cache. Similarly, the Jamba framework \cite{lenz2025jamba} proposes a hybrid design by interleaving standard Transformer layers with Mamba blocks. This approach balances the expressive reasoning capabilities of attention with the linear scaling and high throughput of State Space Models (SSMs) \cite{gu2024mamba}. Furthermore, the InfiniteVL framework \cite{tao2025infinitevl} combines linear and sparse attention to support vision-language modeling with extensive input lengths, achieving notable inference acceleration on standard hardware without increasing the memory footprint. These hybrid architectures demonstrate that the strategic integration of diverse operators is essential for efficient long-range modeling and real-time multimodal deployment on resource-constrained devices.

Despite these advances, existing heterogeneous architectures are subject to two primary limitations that restrict their broader applicability. First, current designs rely on manually designed topological arrangements with fixed mixing ratios, which are inherently suboptimal. These heuristic patterns often employ simple periodic interleaving of different operators, failing to explore the vast architectural search space and reach the optimal performance-efficiency frontier for specific multimodal tasks. Second, these fixed configurations lack the flexibility to adapt to diverse hardware deployment targets. Given that different hardware platforms exhibit distinct execution profiles for various operators, a fixed architecture optimized for one device may fail to meet the strict latency constraints of another. Addressing these hardware-specific requirements through manual trial-and-error is computationally prohibitive, as structural reconfiguration typically necessitates an expensive retraining cycle. Consequently, the inability to automatically customize heterogeneous architectures for specific hardware remains a critical bottleneck for the deployment of LMMs.

To address these challenges, we introduce MOSAIC, which automatically transforms homogeneous VLMs into optimized heterogeneous architectures through adaptive inter-layer composition.
MOSAIC encompasses a unified, per-layer search space that integrates a wide spectrum of efficiency mechanisms, including linear attention, sparse operators, and low-rank representations, allowing for fine-grained architectural adaptation. To identify the optimal configuration without the prohibitive costs of exhaustive end-to-end retraining, we first employ Blockwise Local Distillation (BLD) to initialize candidate modules by aligning them with their corresponding teacher layers. By evaluating candidates across multiple capability metrics, we formulate the architecture selection as a multi-objective MIP problem. This approach enables the discovery of optimal heterogeneous structures that satisfy strict hardware-specific latency constraints while maximizing model performance.
To mitigate the performance degradation in such structural transitions, we introduce a two-stage distillation method. We first perform global off-policy distillation to stabilize the model's internal representations. This is followed by a dual teacher on-policy distillation, which uses a 235B oracle teacher for knowledge expansion and the original 4B teacher to maintain stability. We validate our MOSAIC method through the MOSAIC-4B model, which effectively matches the performance of the Qwen3-VL-4B-Instruct baseline across multiple benchmarks. Notably, it achieves a $1.76\times$ prefilling speedup and $2.54\times$ decoding acceleration, while requiring less than 2\% of the original training cost. The corresponding results are shown in \cref{fig:head_fig}. Our contributions are summarized as follows:

\begin{itemize}[leftmargin=4mm]
\item We propose MOSAIC, which transitions homogeneous VLMs into optimized heterogeneous architectures via adaptive inter-layer composition under hardware-specific latency constraints. To our knowledge, it is the first work to explore the automated search for optimal heterogeneous VLM configurations.
\item We formulate the architectural search as a multi-objective optimization problem, enabling the discovery of heterogeneous mixtures that maximize downstream performance within given hardware constraints.
\item We propose a progressive two-stage parameter recovery method to effectively restore foundational capabilities and achieve high-quality alignment.
\item We demonstrate the effectiveness of the MOSAIC method through the generated MOSAIC-4B model, which achieves significant inference acceleration and matches the overall capabilities of its homogeneous teacher.
\end{itemize}

\section{Related Work}
\label{sec:related}

\noindent\textbf{Efficient Attention and Hybrid Architecture.}
The quadratic computational and memory complexity of standard self-attention \cite{vaswani2017attention} remains the primary obstacle to scaling Large Multimodal Models (LMMs) for long-context or hardware-constrained scenarios. To address this, various efficient attention kernels have been proposed. For instance, Linear Transformers such as RetNet \cite{sun2023retentive} and RWKV \cite{peng2023rwkv} replace the softmax kernel with associative feature maps to enable recurrent inference with $O(N)$ complexity. Similarly, State Space Models (SSMs) like Mamba \cite{gu2024mamba} achieve efficiency through hardware-aware parallel algorithms in recurrent mode. More specialized operators have also been developed. Kimi Delta Attention (KDA) \cite{team2025kimi} employs delta update rules for long-range modeling, Gated Delta Networks (GDN) \cite{yang2024gated} incorporate gating for expressive routing, and Multi-head Latent Attention (MLA) \cite{deepseek2024v2} projects KV pairs into low-rank latent spaces to reduce memory usage. Additionally, Sliding Window Attention (SWA) \cite{child2019generating} and Native Sparse Attention (NSA) \cite{yuan2025native} apply local or dynamic sparsity to balance efficiency and resolution. While these mechanisms reduce complexity, their limited expressivity has historically led to lower performance compared to standard self-attention in language modeling. For example, linear mechanisms such as GDN and KDA achieve comparable performance on moderate-length sequences through gating or decay, but the performance of structures based purely on linear attention remains constrained by finite-state capacity, making long-sequence modeling fundamentally challenging.

Heterogeneous architectures that combine standard self-attention with efficient mechanisms have emerged as a practical approach to balancing quality and efficiency. Jamba \cite{lenz2025jamba} interleaves Transformer layers with Mamba blocks, while InfiniteVL \cite{tao2025infinitevl} and Kimi-Linear \cite{team2025kimi} integrate linear and global attention to process long multimodal inputs. Step 3.5 Flash \cite{huang2026step} combines SWA and GQA to improve decoding efficiency without sacrificing performance. However, these frameworks rely on manually designed, periodic mixing patterns, such as placing one full-attention layer every three layers. These static designs are often suboptimal because they do not account for the different representational needs at various network depths and are difficult to adapt to different hardware. MOSAIC addresses this by automating the orchestration of per-layer mechanisms through a data-driven search.

\noindent\textbf{Neural Architecture Search.}
Neural Architecture Search (NAS) is a key tool for automating the design of efficient models for specific hardware. Early hardware-aware NAS frameworks like HAT \cite{wang2020hat} and OFA \cite{cai2019once} optimized the width and depth of models to meet latency constraints on edge devices. In the context of LMMs, searching for an entirely new architecture and retraining from scratch is often computationally prohibitive. To mitigate this, several works seek to identify architectural design laws at a lower computational cost and extrapolate them to larger models. For instance, \cite{sun2026hardware} establishes a hardware co-design scaling law by modeling training loss as a function of architectural hyperparameters and using roofline modeling to identify the Pareto frontier for inference latency. Similarly, Composer \cite{acun2025composer} performs architecture searches at a small scale before applying specific extrapolation strategies to scale the structures to 3B parameters. Another line of work employs distillation-based search to identify sub-models from a teacher model that satisfy specific constraints. A representative method is Puzzle \cite{bercovich2025puzzle}, which uses the logits of a large teacher to guide the search for an optimal student architecture. Puzzle is closely related to our approach but is typically limited to a search space focused on size, such as reducing the number of layers or hidden dimensions within the same architecture family, and is primarily designed for unimodal text processing. MOSAIC differs by treating the mechanism type including linear, sparse, low-rank, or full attention as a primary search variable. By formulating the architecture search as a multi-objective MIP problem, we identify optimal topologies for VLMs that satisfy strict hardware latency constraints while maximizing downstream performance.

\section{Methodology}
\label{sec:method}

\begin{figure*}[t]
  \centering
  \includegraphics[width=0.99\linewidth]{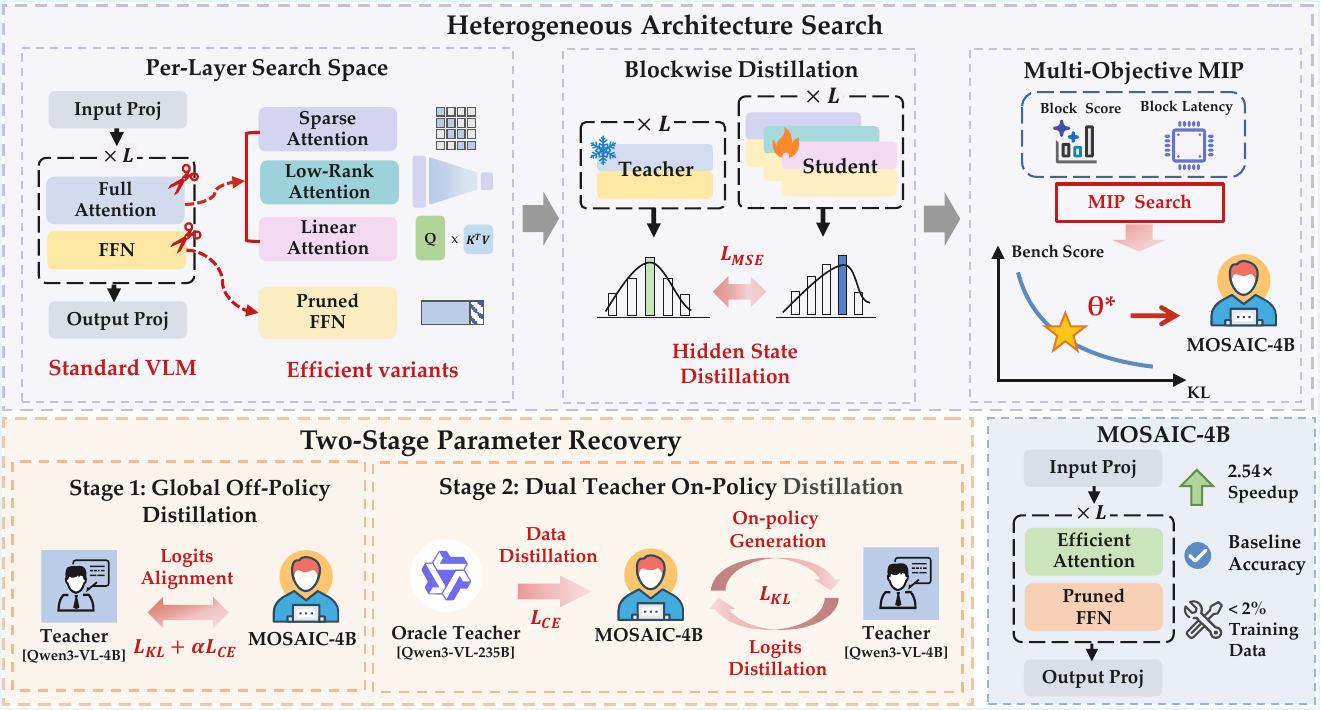}
  \caption{Overview of the MOSAIC method, which consists of two steps: (1) Heterogeneous Architecture Search, which identifies the optimal structure via BLD and multi-objective MIP; (2) Two-Stage Parameter Recovery, which recovers model performance through global off-policy distillation followed by a dual-teacher on-policy distillation. The resulting MOSAIC-4B inherits heterogeneous designs, achieves significant inference acceleration and matches the performance of its homogeneous teacher.}
  \label{fig:overview}
\end{figure*}

In this section, we introduce the proposed MOSAIC method for searching heterogeneous VLM architectures. In \cref{sec:3-1}, we define a unified, per-layer heterogeneous search space that integrates
diverse attention mechanisms and FFN configurations, allowing the model to adapt its architecture across different depths. \cref{sec:3-2} details the search methodology, which involves blockwise local distillation, variant scoring, and the formulation of a multi-objective MIP problem to identify optimal structures under hardware constraints. In \cref{sec:3-3}, we introduce a recovery process to restore model quality through two-stage distillation. An overview of MOSAIC is illustrated in \cref{fig:overview}.

\subsection{Search Space Design}
\label{sec:3-1}

In this work, we define a comprehensive search space for MOSAIC, encompassing diverse architectural operations for each layer of the parent homogeneous VLM. Following the convention of modular design, each transformer layer is treated as a block composed of two primary subblocks: the attention module and the feed-forward network (FFN). The total search space for an $L$-layer model allows each layer to independently select its optimal configuration from a pool of valid candidates. To ensure structural compatibility, following layer-wise NAS techniques \cite{fan2023layernas, bercovich2025puzzle}, we require all candidate subblocks to maintain the same hidden dimensions as the parent model, preserving the consistency of the residual stream.

\textbf{Attention subblocks:} For the attention subblocks in each layer, the search space integrates a spectrum of mechanisms ranging from standard quadratic kernels to efficient linear recurrences and sparse patterns. Specifically, for a teacher model based on Grouped-Query Attention (GQA), the search space includes the GQA configuration of the original teacher model, linear attention mechanisms such as Kimi Delta Attention (KDA) and Gated Delta Network (GDN), the low-rank KV compression mechanism via Multi-head Latent Attention (MLA), and the sparse attention mechanism via Sliding Window Attention (SWA).

\textbf{FFN subblocks:} For the FFN subblocks in each layer, in contrast to the discrete mechanism replacement strategy employed for the attention subblocks, we adopt a structured pruning approach to facilitate more granular utilization of the hardware latency budget. This scaling-centric search space focuses on optimizing the intermediate expansion of the SwiGLU layers through a series of width-reduction configurations. The available choices range from 100\%, which retains the full intermediate dimension of the teacher model for maximum representational capacity, down to 0\%, representing a No-Op configuration that removes the FFN block from the residual stream.

The search space is designed to balance inference latency and model accuracy by addressing the primary computational bottlenecks of standard Transformer architectures. By incorporating KV compression mechanisms such as GQA and MLA, linear attention alternatives such as KDA and GDN, and sparse attention patterns via SWA, the method significantly reduces the time-per-token required for real-time VLM applications. Within this method, FFN scaling serves as a complementary axis of heterogeneity rather than a simple compression step. Because the required representational width of the subsequent FFN may change as attention mechanisms are modified, we incorporate different scaling configurations ranging from 0\% to 100\%. This allows the search process to reallocate the parameter budget from redundant feed-forward blocks to more demanding attention layers, resulting in a more balanced distribution of parameters and computation. Such flexibility enables the model to meet specific hardware acceleration targets while maintaining high performance.

To illustrate the combinatorial complexity of the designed search space, consider the Qwen3-VL-4B-Instruct model, which consists of $L=36$ transformer layers. For the specific instantiation presented in this work, we define five alternatives for attention subblocks and seven scaling configurations for FFN subblocks (from 0\% to 100\%), resulting in 35 possible architectural configurations per layer. Consequently, the total number of potential child model architecture candidates is $35^{36}$. To navigate this large and diverse search space efficiently without the prohibitive cost of end-to-end retraining, we employ the proposed MOSAIC method, comprising blockwise local distillation, blockwise capability scoring, and multi-objective architecture search (detailed in \cref{sec:3-2}).

\subsection{Heterogeneous Architecture Search}
\label{sec:3-2}
The heterogeneous architecture search in MOSAIC consists of three stages: blockwise local distillation, multi-capability scoring, and architecture generation via a multi-objective MIP solver.

\noindent\textbf{Blockwise Local Distillation.}
To provide high-quality initialization for the search space, we first construct a block library by aligning each candidate block with the corresponding layer of the teacher model. We perform Blockwise Local Distillation (BLD), where each candidate block is optimized to match the teacher's output activations at the same layer. Concretely, we minimize the normalized Mean Squared Error (MSE) between the teacher output $\mathbf{o}_t$ and the candidate output $\mathbf{o}_c$:
\begin{equation}
    \mathcal{L}_{\mathrm{align}} = \|\mathbf{o}_t - \mathbf{o}_c\|^2 / \|\mathbf{o}_t\|^2.
\end{equation}
During BLD, each child block receives as input the activations produced by the preceding teacher layer, which isolates the optimization of individual modules and makes the process highly parallelizable. In practice, this stage requires only 80,000 training samples to establish a strong initialization for the entire candidate pool. More importantly, it preserves compatibility with the original residual stream, thereby reducing the difficulty of the subsequent global recovery stage.

\noindent\textbf{Multi-Capability Scoring.}
Although BLD enforces local alignment, it does not fully reflect the global generalization ability of a candidate block or its effect on challenging downstream tasks. We therefore introduce an ability set $\Lambda$ to evaluate each candidate from multiple perspectives, including Perplexity (PPL), Kullback--Leibler (KL) divergence, and representative LLM/VLM benchmark scores. Compared with prior methods that rely primarily on KL divergence as a proxy for alignment quality \cite{bercovich2025puzzle}, our formulation incorporates both language modeling quality and task-level performance, leading to a more reliable estimate of each block's capability.

For the $i$-th candidate block at layer $l$, we define a relative capability score $S^k_{l,i}$ for the $k$-th ability in $\Lambda$. Following the normalization practice commonly used in Multi-Task Learning (MTL) \cite{zhang2021survey, linreasonable}, we compute this score as the normalized improvement over a reference baseline:
\begin{equation}
    S_{l,i}^k = (-1)^{\sigma_k} \cdot \frac{P_{l,i}^k - M^{k}}{M^{k}},
\end{equation}
where $P_{l,i}^k$ denotes the performance achieved when the $i$-th candidate is inserted into layer $l$, and $M^{k}$ is the corresponding reference value. For PPL and downstream benchmarks, $M^{k}$ is defined as the teacher's performance; for KL divergence, it is set to the minimum KL value among all candidates across all layers. The binary variable $\sigma_k$ indicates the direction of the metric: $\sigma_k = 1$ for ``lower-is-better'' metrics such as PPL and KL, and $\sigma_k = 0$ for ``higher-is-better'' metrics such as benchmark accuracy. Under this formulation, a positive $S_{l,i}^k$ always indicates a favorable candidate, which makes scores from heterogeneous objectives directly comparable during optimization.

\noindent\textbf{Multi-Objective MIP Formulation.}
After obtaining the per-block capability scores and hardware runtime profiles, we formulate architecture selection as a multi-objective constrained optimization problem. The goal is to identify a heterogeneous configuration that satisfies a predefined latency budget while preserving the strongest and most balanced capability profile across the ability set $\Lambda$.

Consider a teacher model with $L$ layers and $K$ candidate blocks per layer. For the $i$-th candidate at layer $l$, we introduce a binary decision variable $x_{l,i} \in \{0,1\}$ indicating whether that candidate is selected. The architecture configuration $\mathbf{X}$ is therefore defined as
\begin{equation}
    \mathbf{X} = \{ x_{l,i} \mid 1\le l\le L, 1\le i \le K\} \in \mathcal{B},
\end{equation}
where $\mathcal{B}$ denotes the feasible set satisfying $\sum_{i=1}^K x_{l,i} = 1$ for every layer and $x_{l,i} \in \{0,1\}$.

Assuming that $m$ abilities are considered, the architecture-level score on the $k$-th ability is defined as
\begin{equation}
    F^k(\mathbf{X}) = \sum\nolimits_{i=1}^K \sum\nolimits_{l=1}^L x_{l,i} S^k_{l,i}.
\end{equation}
The resulting multi-objective optimization problem is written as
\begin{align}
\max_{\mathbf{X}\in \mathcal{B}} \quad & \Big(F^1(\mathbf{X}), F^2(\mathbf{X}), \ldots, F^m(\mathbf{X})\Big), \label{eq:mo-obj}\\
\text{s.t.} \quad
& \sum\nolimits_{i=1}^K \sum\nolimits_{l=1}^L \mathsf{runtime}(l,i)\,x_{l,i} \leq \mathsf{Target},
\end{align}
where $\mathsf{runtime}(l,i)$ is the measured runtime of candidate $i$ at layer $l$, and $\mathsf{Target}$ is the global latency budget corresponding to the desired acceleration ratio. This formulation explicitly captures the trade-off between efficiency and capability retention.

To solve the problem, we adopt the Tchebycheff (TCH) scalarisation \cite{lin2024smooth}, which can recover Pareto-optimal solutions even on non-convex regions of the Pareto frontier. Given a preference coefficient $\mathbf{w} = (w^1, \ldots, w^m)$ with $\sum_k w^k = 1$ and an ideal point $\mathbf{z} = (z^1, \ldots, z^m)$ where $z^k = \max_{\mathbf{X}\in \mathcal{B}} F^k(\mathbf{X})$, the TCH objective becomes
\begin{align}\label{eq:min_max}
\min_{\mathbf{X}\in \mathcal{B}} \max_{k=1, \ldots, m} \quad & w^k \cdot \left(z^k - F^k(\mathbf{X})\right), \\
\text{s.t.} \quad
& \sum\nolimits_{i=1}^K \sum\nolimits_{l=1}^L \mathsf{runtime}(l,i)\,x_{l,i} \leq \mathsf{Target}.
\end{align}
This min--max objective can be converted into a standard linear MIP by introducing an auxiliary variable $C$:
\begin{align}
\min_{\mathbf{X}\in \mathcal{B}} \quad & C, \label{eq:tch-mip}\\
\text{s.t.} \quad
& w^k \cdot \left(z^k - F^k(\mathbf{X})\right) \leq C, \quad k = 1, \ldots, m, \\
& \sum\nolimits_{i=1}^K \sum\nolimits_{l=1}^L \mathsf{runtime}(l,i)\,x_{l,i} \leq \mathsf{Target}.
\end{align}
The resulting problem is solved efficiently with the open-source \texttt{python-mip} package \cite{python-mip}. In our experiments, we consider $m=4$ objectives in $\Lambda$: KL divergence, PPL, LLM benchmark performance, and VLM benchmark performance. Together, these objectives provide a balanced criterion that spans local alignment, language modeling quality, and downstream multimodal capability.

\subsection{Two-Stage Parameter Recovery}
\label{sec:3-3}
A key limitation of blockwise local distillation is the lack of inter-block compatibility, as modules are trained using teacher inputs rather than the actual outputs of their predecessors in the student model. Therefore, the searched model initially exhibits a performance gap. To bridge this gap and restore the model's capabilities, we introduce a global knowledge distillation (GKD) phase as the final step in the proposed MOSAIC method, which contains a two-stage training pipeline: \textit{Global Off-Policy Distillation} and \textit{Dual Teacher On-Policy Distillation}.

\noindent\textbf{Global Off-Policy Distillation.} The first stage focuses on stabilizing the internal representations of the student by distilling knowledge from the given homogeneous teacher model (\eg, Qwen3-VL-4B-Instruct). We utilize a diverse training set comprising large-scale text corpora to recover language modeling proficiency and interleaved image-video-text pairs to restore cross-modal alignment.

During this stage, the student is trained to replicate the output distribution of the teacher through a standard knowledge distillation objective comprising a hard-target loss and a soft-target loss. We denote the probability distributions from the teacher and student model by $p$ and $q_{\theta}$, respectively, where $\theta$ is the model parameter.
The corresponding hard-target loss is denoted by $\mathcal{L}_{\mathrm{CE}}(q_{\theta};\mathcal{D})$, which represents the Cross-Entropy (CE) loss calculated on the ground-truth tokens in datasets $\mathcal{D}$. For the soft-target distillation, following \cite{sessa2024bond, wu2025rethinking}, we adopt a symmetric Kullback-Leibler Divergence, which combines forward KL and reverse KL divergences, and is denoted by
\begin{equation}
    \mathcal{L}_{\mathrm{KL}}(\theta;\mathcal{D})= \mathrm{D}_\mathrm{KL}\!\left(p\| q_{\theta}\right) + \mathrm{D}_\mathrm{KL}\!\left(q_{\theta}\| p\right).
\end{equation}
The forward KL term encourages the student to assign probability mass where the teacher does, ensuring broad coverage of the teacher's knowledge. Conversely, the reverse KL term penalizes the student for placing mass where the teacher does not, yielding sharper and more confident predictions. This combination provides complementary supervision, balancing broad distributional coverage with focused sharpening.

The overall training objective for global off-policy distillation is defined as:
\begin{equation}
    \mathcal{L}_{\mathrm{recover}}(\theta) =  \mathcal{L}_{\mathrm{KL}}(\theta;\mathcal{D}) + \alpha\mathcal{L}_{\mathrm{CE}}(q_{\theta};\mathcal{D}),
\end{equation}
where $\alpha$ is a hyperparameter. This global off-policy distillation provides a robust initialization, effectively recovering the general reasoning and perception capabilities of the teacher model.

\noindent\textbf{Dual Teacher On-Policy Distillation.}
While Global Off-Policy Distillation already aligns the student with the baseline teacher on foundational abilities, further training is still required for the student to improve on complex reasoning and challenging tasks. To push the performance ceiling without inducing catastrophic forgetting or optimization instability, we propose a dual-teacher instruction alignment strategy. This stage simultaneously leverages a large-capacity oracle teacher (\eg, Qwen3-VL-235B) for knowledge expansion and the original teacher for distributional stability.

To instill superior reasoning patterns beyond the capacity of the baseline teacher, we utilize responses synthesized by the oracle teacher as high-quality targets. We apply a standard Supervised Fine-Tuning (SFT) loss on these outputs, allowing the student to learn what to generate from a more powerful policy. This prevents the student from being limited by the inherent architectural constraints of the baseline teacher.

Furthermore, to ensure training stability and prevent the student from drifting away from its functional distribution, we employ the baseline teacher model as an on-policy logits teacher. For each prompt, the student generates a response based on its current policy; the baseline teacher then provides supervision by computing logits for these student-generated tokens. We apply the mixed KL divergence as the on-policy distillation loss, ensuring the student learns how to maintain a stable output distribution while absorbing new knowledge from the oracle teacher.

The final objective for the dual teacher on-policy distillation is formulated as:
\begin{equation}
    \mathcal{L}_{\mathrm{align}}(\theta) =  \mathcal{L}_{\mathrm{KL}}(\theta;\mathcal{D}')+\beta\mathcal{L}_{\mathrm{CE}}(\theta; \mathcal{D}_{o}),
\end{equation}
where $\mathcal{D}_{o}$ represents the data synthesized by the large-capacity oracle teacher, and $\mathcal{D}'$ denotes the on-policy supervision from the baseline teacher model.
This hybrid approach allows the student to surpass the performance of the baseline teacher while remaining grounded in a stable optimization landscape.

\section{Experiments}
\label{sec:exp}
In this section, we evaluate the effectiveness of MOSAIC by deriving heterogeneous VLM models and conducting comprehensive ablation studies.

\subsection{Experimental Setup}
\label{sec:setup}
We apply the MOSAIC method to Qwen3-VL-4B-Instruct, a state-of-the-art multimodal model supporting both text and multimodal understanding. Through this process, we generate MOSAIC-4B, an optimized heterogeneous derivative.

\noindent\textbf{Searching configuration.}
The search space comprises 35 valid candidates per layer across $L=36$ transformer layers, yielding a combinatorial space of size $35 \times 36 = 1260$ for the MOSAIC method.
Following \cite{lin2024smooth}, the weight vector $\omega$ is assigned equal preference across all objectives.
The runtime of each block variant is evaluated at batch size 1 with a prefilling length of 96k and a decoding length of 256, where the corresponding results for all block variants are shown in \cref{sec:appendix_b1}. The predefined global latency budget is set with a constraint of $1.5\times$ speedup w.r.t.\ Qwen3-VL-4B-Instruct baseline model. For the capability scoring, we utilize both plain-text and multimodal inputs: 1K long-text instances sampled from Olmo3~\cite{olmo2025olmo} for PPL and KL divergence, MMLU~\cite{hendrycks2020measuring} for evaluating plain-text capability, and SeedBench~\cite{li2024seed} for evaluating multimodal capability.

\noindent\textbf{Training configuration.} \textit{Due to the
page limitation, the training details are put in \cref{sec:appendix_b2}.} The whole training corpus comprises exclusively publicly available, open-source datasets totaling $\sim$32M samples. Therefore, the total training cost is less than 2\% training cost of the original Qwen3-VL-4B-Instruct.

\noindent\textbf{Evaluation benchmarks.}
We evaluate the generated MOSAIC-4B model and baseline models on 19 representative benchmarks spanning three modalities.
\emph{Image understanding} (13 benchmarks): MME~\cite{fu2023mme}, MMBench-EN, MMBench-CN~\cite{liu2024mmbench}, MMMU~\cite{yue2024mmmu}, GQA~\cite{hudson2019gqa}, ScienceQA~\cite{lu2022learn}, BLINK~\cite{fu2024blink}, CV-Bench 2D, CV-Bench 3D~\cite{tong2024cambrian}, InfoVQA~\cite{mathew2022infographicvqa}, DocVQA~\cite{mathew2021docvqa}, ChartQA~\cite{masry2022chartqa}, and OCRBench~\cite{mishra2019ocr}.
\emph{Video understanding} (3 benchmarks): MVBench~\cite{li2024mvbench}, Video-MME~\cite{fu2025video}, and WorldSense~\cite{benchekroun2023worldsense}.
\emph{Text understanding} (3 benchmarks): ARC-Challenge~\cite{clark2018think}, GSM-8K~\cite{cobbe2021training}, and HellaSwag~\cite{zellers2019hellaswag}.

\subsection{Main Results}
\label{sec:main_results}
The performance of the proposed MOSAIC-4B model on image understanding, video understanding, and text understanding are reported on \cref{tab:main_image}, \cref{tab:main_video}, and \cref{tab:main_text}, respectively. The Qwen3-VL-4B-Instruct model is used as the reference model, and we report the average performance and average per-metric difference, denoted by $\Delta$, for every competing baseline model.

\begin{table*}[t]
  \caption{\textbf{Image understanding results and the time per output token (TPOT) speedup.}
  \colorbox{gray!15}{Gray row}: Qwen3-VL-4B-Instruct reference.
  $\Delta$: mean per-metric gap (\%) w.r.t.\ the reference.
  Avg: average performance (\%) over all 13 metrics.
  \textbf{Bold}: best; ${}^\dagger$: second best among competing models.}
  \label{tab:main_image}
  \centering
  \setlength{\tabcolsep}{3pt}
  \resizebox{\linewidth}{!}{%
  \begin{tabular}{@{}lc|*{13}{c}|cc@{}}
    \toprule
    Model & TPOT
      & MME & MMB-E & MMB-C & MMMU & GQA & SQA & BLINK
      & CV-2D & CV-3D & Info & Doc & Chart & OCR
      & Avg & $\Delta$\,$\uparrow$ \\
    \midrule
    \rowcolor{gray!15}
    Qwen3-VL-4B-Instruct~\cite{bai2025qwen3} &
     $1.00\times$ & 2299 & 83.7 & 76.1 & 61.5 & 62.9 & 88.6 & 65.1
      & 69.9 & 88.5 & 79.9 & 94.2 & 84.2 & 84.1
      & 78.5 & 0.0 \\
    \midrule
    InternVL3.5-4B~\cite{wang2025internvl3} &
      $1.41\times$ & 2272 & 80.3 & \textbf{79.5} & \textbf{66.6} & $62.4^\dagger$ & $86.3^\dagger$ & $58.4^\dagger$
      & \textbf{72.3} & 85.2 & \textbf{78.4} & $93.2^\dagger$ & \textbf{86.4} & $82.2^\dagger$
      & $77.9^\dagger$ & $-0.6^\dagger$ \\
    InternVL3.5-2B~\cite{wang2025internvl3} &
      $2.11\times$ & 2129 & 78.2 & 76.5 & 51.8 & 56.6 & 68.3 & 51.4
      & 58.3 & 80.2 & 69.3 & 70.5 & 63.8 & 63.4
      & 66.5 & $-$12.0 \\
    Qwen3-VL-2B-Instruct~\cite{bai2025qwen3} &
      $2.09\times$ & $2215^\dagger$ & $80.5^\dagger$ & $78.2^\dagger$ & 58.3 & 60.8 & 78.5 & 56.8
      & 64.2 & 83.6 & 73.8 & 82.5 & 76.3 & 73.5
      & 72.6 & $-$5.9 \\
    SmolVLM2-2.2B~\cite{marafioti2025smolvlm} &
      $1.24\times$ & 1892 & 69.2 & 55.3 & 41.6 & 52.4 & 65.1 & 42.3
      & 45.2 & 60.3 & 37.8 & 45.3 & 42.6 & 62.5
      & 52.9 & $-$25.6 \\
    Molmo2-4B~\cite{clark2026molmo2} &
      $0.87\times$ & 2100 & 78.3 & 68.5 & 54.2 & 53.4 & 76.2 & 50.6
      & 66.3 & $85.4^\dagger$ & 72.5 & 84.3 & 75.2 & 70.4
      & 70.0 & $-$8.5 \\
    \textbf{MOSAIC-4B (Ours)} &
      $\mathbf{2.54\times}$ & \textbf{2285} & \textbf{82.7} & 77.6 & $59.9^\dagger$ & \textbf{62.9} & \textbf{89.2} & \textbf{63.7}
      & $70.6^\dagger$ & \textbf{89.3} & 74.6 & \textbf{93.9} & $83.2^\dagger$ & \textbf{83.9}
      & \textbf{77.9} & $\mathbf{-0.6}$ \\
    \bottomrule
  \end{tabular}}
\end{table*}

\begin{table}[t]
  \caption{\textbf{Video understanding results.}
  The notations follow \cref{tab:main_image}.}
  \label{tab:main_video}
  \centering
  \setlength{\tabcolsep}{4pt}
  \resizebox{\linewidth}{!}{%
  \begin{tabular}{@{}l*{4}{c}c@{}}
    \toprule
    Model & MVBench & VideoMME & WorldSense & Avg
      & $\Delta$\,$\uparrow$ \\
    \midrule
    \rowcolor{gray!15}
    Qwen3-VL-4B-Instruct
      & 67.1 & 64.8 & 39.6 & 57.2 & 0.0 \\
    \midrule
    InternVL3.5-4B
      & $66.4^\dagger$ & $63.8^\dagger$ & $38.5^\dagger$ & $56.2^\dagger$ & $-1.0^\dagger$ \\
    InternVL3.5-2B
      & 61.5 & 58.4 & 34.0 & 51.3 & $-$5.9 \\
    Qwen3-VL-2B-Instruct
      & 64.8 & 62.5 & 37.2 & 54.8 & $-$2.4 \\
    SmolVLM2-2.2B
      & 46.3 & 52.1 & 24.5 & 41.0 & $-$16.2 \\
    Molmo2-4B
      & \textbf{69.3} & \textbf{64.5} & 38.2 & \textbf{57.3}
      & $\mathbf{+0.1}$ \\
    \textbf{MOSAIC-4B (Ours)}
      & 66.2 & 63.3 & \textbf{39.7} & $56.4^\dagger$
      & $-0.8^\dagger$ \\
    \bottomrule
  \end{tabular}}
\end{table}

\begin{table}[t]
  \caption{\textbf{Text understanding results.}
  Notation follows \cref{tab:main_image}.}
  \label{tab:main_text}
  \centering
  \setlength{\tabcolsep}{4pt}
  \resizebox{\linewidth}{!}{%
  \begin{tabular}{@{}l*{4}{c}c@{}}
    \toprule
    Model & GSM-8K & ARC & HellaSwag & Avg
      & $\Delta$\,$\uparrow$ \\
    \midrule
    \rowcolor{gray!15}
    Qwen3-VL-4B-Instruct
      & 82.3 & 88.7 & 78.4 & 83.1 & 0.0 \\
    \midrule
    InternVL3.5-4B
      & \textbf{81.2} & \textbf{87.9} & \textbf{77.5} & \textbf{82.2}
      & $\mathbf{-0.9}$ \\
    InternVL3.5-2B
      & 76.5 & 83.2 & 74.1 & 77.9 & $-$5.2 \\
    Qwen3-VL-2B-Instruct
      & 79.5 & $86.2^\dagger$ & 76.3 & 80.7 & $-$2.4 \\
    SmolVLM2-2.2B
      & 43.8 & 56.4 & 54.2 & 51.5 & $-$31.6 \\
    Molmo2-4B
      & 68.5 & 75.4 & 69.2 & 71.0 & $-$12.1 \\
    \textbf{MOSAIC-4B (Ours)}
      & $79.6^\dagger$ & 85.7 & $77.1^\dagger$ & $80.8^\dagger$
      & $-2.3^\dagger$ \\
    \bottomrule
  \end{tabular}}
\end{table}

On image understanding tasks, MOSAIC-4B leads on 8 of 13 metrics with a mean delta of only $-0.6\%$ from the Qwen3-VL-4B-Instruct reference. Notably, it significantly outperforms the Qwen3-VL-2B-Instruct and InternVL3.5-2B baselines, demonstrating that MOSAIC effectively preserves general perception and document comprehension even with reduced latency. On video understanding tasks, Molmo2-4B is the only model to match the reference on average with a $\Delta$ of $+0.1\%$, while MOSAIC-4B obtains the highest WorldSense score of 39.7\% and a competitive $\Delta$ of $-0.8\%$. It also shows a clear margin over the 2B-parameter models, indicating that our heterogeneous allocation is superior to simple parameter reduction for temporal modeling. On text tasks, MOSAIC-4B ranks second at a $\Delta$ of $-2.3\%$. Despite the aggressive speedup, MOSAIC-4B maintains a substantial lead over Qwen3-VL-2B-Instruct and InternVL3.5-2B, confirming that heterogeneous mixing preserves strong language ability alongside visual competence.

\begin{figure}[t]
  \centering
  \includegraphics[width=\columnwidth]{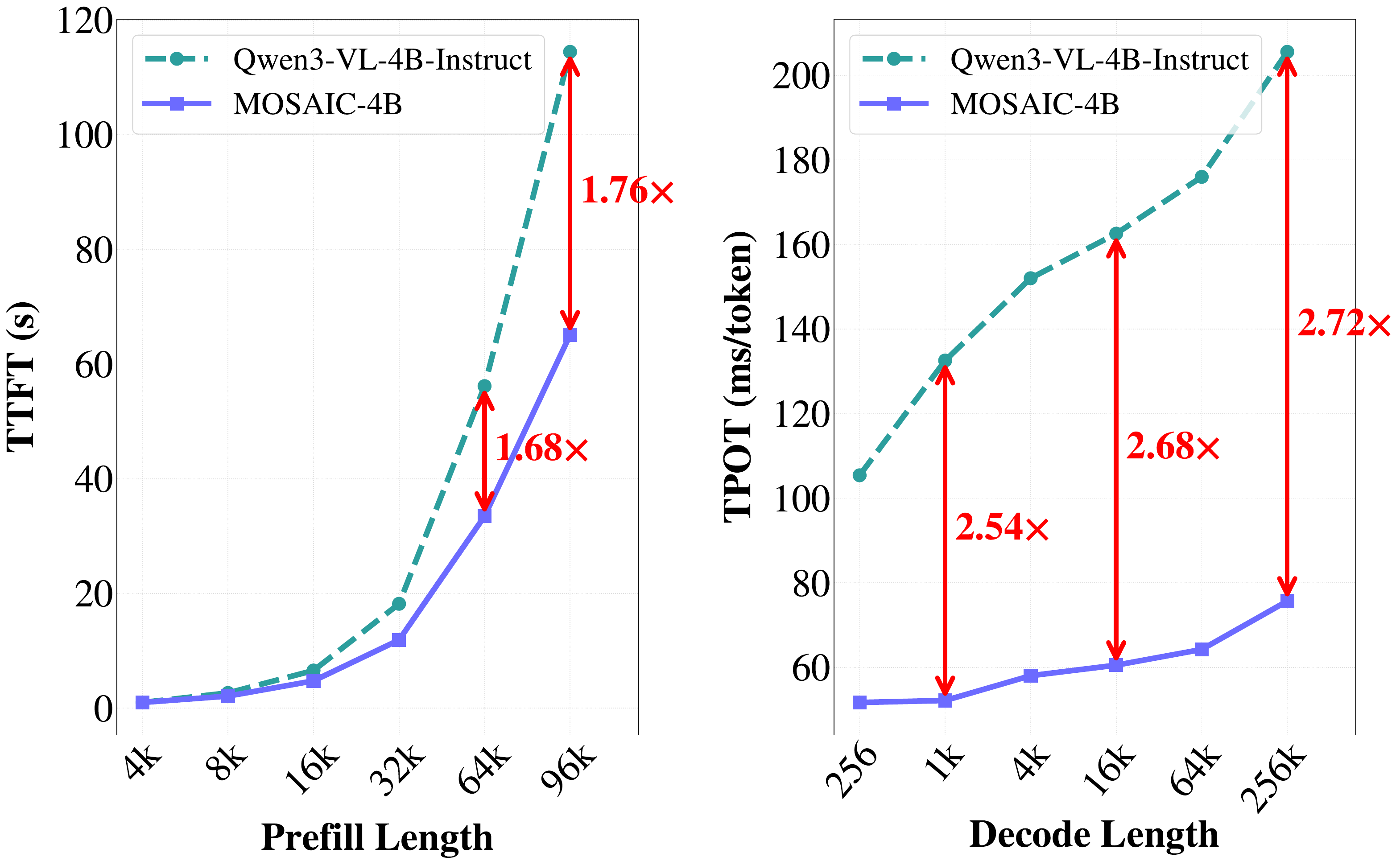}
  \caption{Left: The time to first token (TTFT) of Qwen3-VL-4B-Instruct and MOSAIC. Right: The time per output token (TPOT) of Qwen3-VL-4B-Instruct and MOSAIC, with the prefix length fixed at 96k tokens.}
  \label{fig:ttft_tpot}
\end{figure}

To evaluate practical deployability, we implement MOSAIC-4B within an optimized vLLM framework~\cite{kwon2023efficient}, as detailed in \cref{sec:appendix_b4}. Through this implementation, MOSAIC-4B can run efficiently in real inference scenarios. We evaluate its latency gains w.r.t.\ Qwen3-VL-4B-Instruct model, and the results are illustrated in \cref{fig:ttft_tpot}. Specifically, for prefilling, the speedup in time to first token (TTFT) becomes more substantial with longer prefill lengths, reaching $1.76\times$ at 96k tokens. For decoding, the speedup in time per output token (TPOT) also increases with decode length, attaining $2.54\times$, $2.68\times$, and $2.72\times$ at decode lengths of 1k, 16k, and 256k, respectively. These results indicate that the efficiency advantages of MOSAIC are amplified in long-context and long-generation regimes, where latency is most critical in practice.

\subsection{Ablation Studies}
\label{sec:ablation}
Ablation experiments are performed using 20\% of the training data used for the main model, while other settings remain unchanged.

\subsubsection{Hand-Designed Fixed-Ratio Architectures.}
\label{sec:ablation_ratio}

To evaluate the benefits of automated architectural optimization, we compare MOSAIC against six hand-designed fixed-ratio configurations. Although fixed-ratio interleaving also introduces a form of per-layer heterogeneity, these static patterns are typically designed through manual intuition rather than data-driven search. Our baselines include three configurations from previous studies \cite{team2025kimi, tao2025infinitevl, huang2026step} and three derived from common design patterns, all employing a uniform 1:3 interleaving where every fourth layer utilizes a minority type. For each configuration, we use MOSAIC to search for an optimal per-layer assignment within the same mechanism palette and latency budget. All resulting models are trained using the identical recovery pipeline to ensure that performance gains are attributable solely to the discovered architectural patterns.

\begin{table}[t]
  \caption{\textbf{Hand-designed vs.\ MOSAIC with fixed-ratio architecture configuration.}
  Each group pairs a uniform 1:3 interleaving pattern with a MOSAIC-optimised
  assignment using the same two mechanism types and the same latency budget.
  Design references:
  \textsuperscript{\dag}Kimi,
  \textsuperscript{\ddag}InfiniteVL,
  \textsuperscript{\S}Step-Fun.}
  \label{tab:ablation_ratio}
  \centering
  \setlength{\tabcolsep}{6pt}
  \resizebox{\columnwidth}{!}{%
  \begin{tabular}{@{}llcccc@{}}
    \toprule
    Configuration & Method & Img & Vid & Txt & Overall \\
    \midrule
    \multirow{2}{*}{MLA:KDA = 1:3\textsuperscript{\dag}}
      & Designed    & 57.4 & 39.6 & 49.2 & 48.7 \\
      & MOSAIC & 58.3 & 40.1 & 50.3 & \textbf{49.6} \\
    \midrule
    \multirow{2}{*}{SWA:GDN = 1:3\textsuperscript{\ddag}}
      & Designed    & 64.3 & 41.6 & 62.3 & 56.1 \\
      & MOSAIC & 66.8 & 45.3 & 69.7 & \textbf{60.6} \\
    \midrule
    \multirow{2}{*}{GQA:SWA = 1:3\textsuperscript{\S}}
      & Designed    & 70.6 & 29.6 & 66.3 & 55.5 \\
      & MOSAIC & 72.6 & 31.3 & 64.1 & \textbf{56.0} \\
    \midrule
    \multirow{2}{*}{GQA:GDN = 1:3}
      & Designed    & 71.6 & 44.1 & 61.1 & 58.9 \\
      & MOSAIC & 72.9 & 45.2 & 59.7 & \textbf{59.3} \\
    \midrule
    \multirow{2}{*}{GQA:MLA = 1:3}
      & Designed    & 67.9 & 40.9 & 59.7 & 56.2 \\
      & MOSAIC & 68.7 & 48.7 & 66.4 & \textbf{61.3} \\
    \midrule
    \multirow{2}{*}{GQA:KDA = 1:3}
      & Designed    & 69.8 & 46.9 & 61.6 & 59.4 \\
      & MOSAIC & 70.0 & 52.3 & 63.9 & \textbf{62.1} \\
    \bottomrule
  \end{tabular}}
\end{table}

\cref{tab:ablation_ratio} reveals several findings.
\textbf{(i) Search consistently outperforms fixed ratios.}
Across all six mechanism pairs, the MOSAIC-optimised assignment surpasses the uniform
1:3 interleaving, confirming that different layers exhibit different sensitivity to
attention mechanism choice and that a one-size-fits-all ratio fails to achieve optimal performance.
\textbf{(ii) The choice of mechanism palette matters.}
Even with optimal per-layer placement, some two-mechanism combinations yield
substantially higher ceilings than others,
highlighting the importance of a \emph{rich} search space rather than merely a
smarter allocation of a narrow palette.

\subsubsection{Search Space Design.}
\label{sec:ablation_space}

\begin{table}[t]
  \caption{\textbf{Search space ablation.}
  Each row removes one attention or FFN category from the search space.
  All models are re-searched and re-trained under the same budget.}
  \label{tab:ablation_space}
  \centering
  \setlength{\tabcolsep}{8pt}
  \resizebox{\columnwidth}{!}{%
  \begin{tabular}{@{}lcccc@{}}
    \toprule
    Search Space Variant & Img & Vid & Txt & Overall \\
    \midrule
    Full MOSAIC                    & 74.8 & \textbf{53.6} & \textbf{71.6} & \textbf{66.7} \\
    \midrule
    w/o Linear (KDA, GDN)        & 72.3 & 16.9 & 65.8 & 51.7 \\
    w/o Low-Rank (MLA)     & \textbf{75.0} & 51.6 & 69.8 & 65.5 \\
    w/o Sparse (SWA)             & 73.6 & 52.3 & 69.2 & 65.0 \\
    w/o FFN Search               &  74.6  &   52.9 &  71.3  & 66.3 \\
    \bottomrule
  \end{tabular}}
\end{table}

To assess the importance of different architectural components, we ablate the search space by removing one mechanism family at a time while maintaining a fixed latency budget of $1.5\times$ speedup relative to the Qwen3-VL-4B-Instruct baseline. The results reported in \cref{tab:ablation_space} reveal that linear attention variants are the most critical component. Removing KDA and GDN results in a substantial overall drop of 15.0, a decline driven almost entirely by the video modality, where performance falls from 53.6\% to 16.9\%. This occurs because the solver is forced to replace linear layers that are effective for long contexts with costlier alternatives or layer removal to satisfy the latency constraints. Sparse attention is the second most impactful with a reduction of 1.7\%, followed by low-rank attention with a reduction of 1.2\%. While both incur moderate degradation across all modalities, the impact remains far less severe than the loss of linear attention. These results confirm that maintaining a diverse attention set, particularly linear attention mechanisms, for long sequences, is essential to the effectiveness of the MOSAIC method.

\subsubsection{Multi-Objective Architecture Search.}
\label{sec:ablation_mo}

\begin{table}[t]
  \caption{\textbf{Ablation on search objectives.}
  Objectives are added incrementally.
  All variants use the same latency budget and recovery pipeline.}
  \label{tab:ablation_mo}
  \centering
  \setlength{\tabcolsep}{8pt}
  \begin{tabular}{@{}lcccc@{}}
    \toprule
    Search Objectives & Img   & Vid   & Txt            & Overall  \\
    \midrule
    KL only               & 65.3  & 42.9  & 71.2           & 59.8 \\
    + PPL             & 66.7  & 42.3  & 70.3           & 59.8 \\
    + LLM Bench       & 69.7  & 48.8  & \textbf{71.9}  & 63.5 \\
    + VL Bench        & \textbf{74.7} & \textbf{53.2} & 70.1 & \textbf{66.0} \\
    \bottomrule
  \end{tabular}
\end{table}

We conduct an ablation study on the effectiveness of multi-objective architecture search. As we claimed in \cref{sec:3-2}, the KL divergence can be considered as an indicator of alignment quality. Therefore, starting from a single objective architecture search based on KL divergence, we sequentially incorporate objectives, including PPL, LLM benchmark performance, and VLM benchmark performance.

The corresponding results are shown in \cref{tab:ablation_mo}. The searched model based on KL divergence alone yields the weakest overall performance of 59.8\%. This suggests that minimizing distributional divergence on calibration data does not directly translate to strong task-level accuracy. Introducing PPL leaves the overall score unchanged at 59.8\% and produces only marginal shifts across modalities, which indicates that perplexity alone provides limited complementary guidance. The LLM benchmark objective brings the largest single-step improvement of 3.7\% points overall. This substantially boosts video and image performance by 6.5\% and 3.0\%, respectively, while retaining the best text score of 71.9\%. These results demonstrate the importance of steering the search toward reasoning-preserving configurations. Finally, incorporating the VL benchmark objective delivers further gains on image at 74.7\% and video at 53.2\%, lifting the overall score to 66.0\% at only a modest text trade-off of 1.8\%. The full four-objective formulation achieves the best overall balance, confirming that each additional objective contributes non-redundant, modality-aware information to the architecture search.

\subsubsection{Two-Stage Parameter Recovery.}
\label{sec:ablation_recovery}

\begin{table}[t]
  \caption{Ablation on the parameter recovery pipeline.
  Performance impact of different training stages and objectives. Stage 2 variants are applied after Stage 1.}
  \label{tab:ablation_recovery}
  \centering
  \setlength{\tabcolsep}{6pt}
  \resizebox{\columnwidth}{!}{%
  \begin{tabular}{@{}lcccc@{}}
    \toprule
    Configuration & Img & Vid & Txt & Overall \\
    \midrule
    Stage 1: Pre-train KD           & 60.6 & 45.6 & 61.8 & 56.0 \\
    Stage 1: Instruction KD         & 76.0 & 53.3 & 74.9 & 68.1 \\
    \midrule
    Stage 2: SFT only   & 72.9 & 44.9 & 71.6 & 63.1 \\
    Stage 2: On-policy KD only & 76.1 & 54.3 & 75.9 & 68.8 \\
    Stage 2: SFT + On-policy KD & \textbf{77.9} & \textbf{56.4} & \textbf{80.8} & \textbf{71.7} \\
    \bottomrule
  \end{tabular}}
\end{table}

To evaluate the contribution of each recovery component, we progressively add elements of the two-stage pipeline, with the results in \cref{tab:ablation_recovery} showing a clear progression. In Stage 1, instruction KD yields a substantial overall improvement of 12.1\% on top of pre-training KD, raising the performance from 56.0\% to 68.1\%. This stage includes significant gains across all modalities, such as an increase of 15.4\% on image understanding, which confirms that task-specific instruction alignment is essential. In Stage 2, applying SFT alone leads to a performance drop from 68.1\% to 63.1\%, likely due to distribution shift or catastrophic forgetting. Conversely, applying on-policy KD alone provides a slight overall gain of 0.7\% while maintaining stability. The full pipeline, combining both SFT and on-policy KD, yields the best result of 71.7\%, representing a 3.6\% improvement over Stage 1 alone. This confirms that the supervised data generated by 235B oracle teacher model provides complementary knowledge, but this signal must be regularized by on-policy KD to maintain distributional consistency and achieve optimal performance.

\subsection{Discovered Architecture Patterns}
\label{sec:discover}

\begin{figure}[t]
  \centering
  \includegraphics[width=\columnwidth]{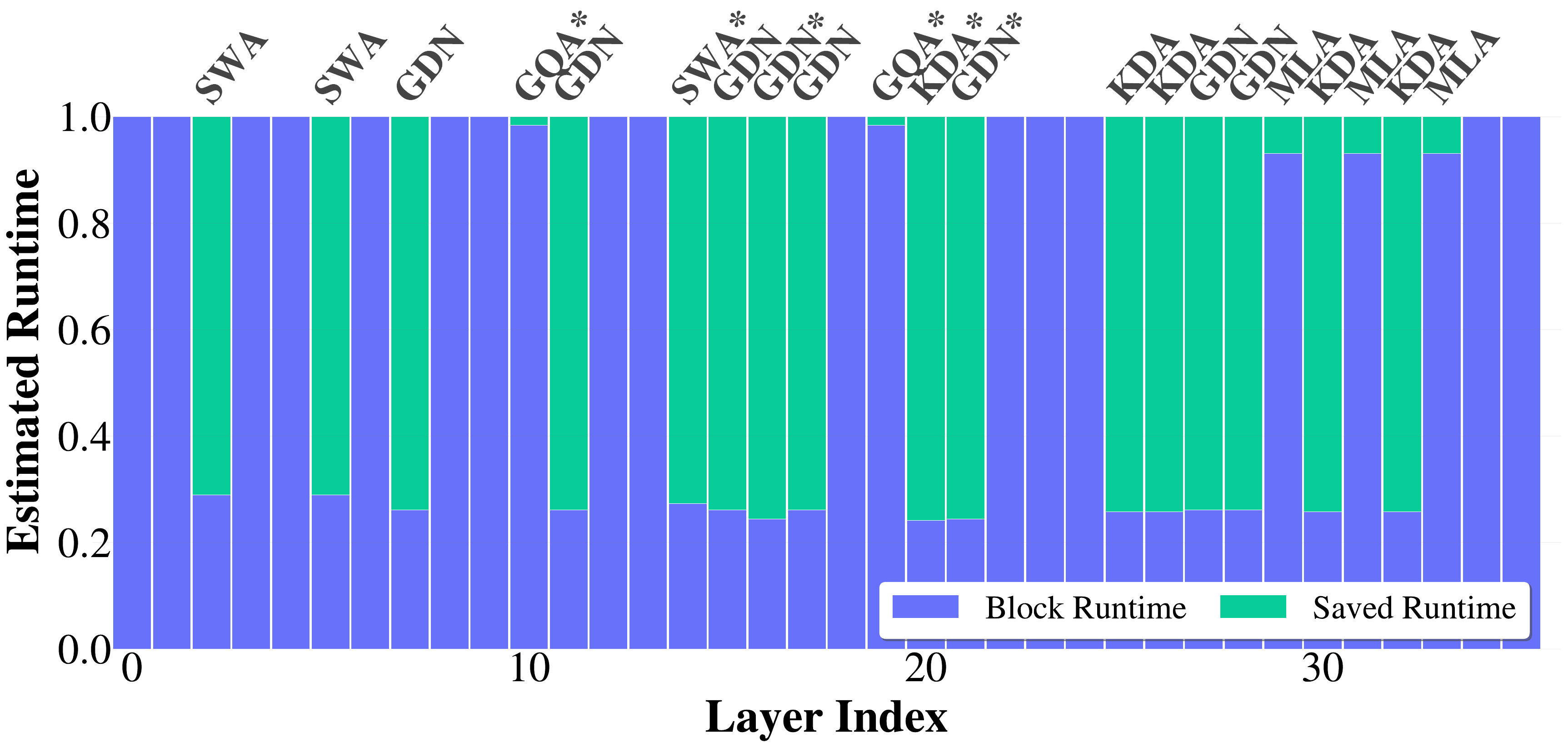}
  \caption{The runtime reduction of layer-wise architectural replacements in MOSAIC, relative to the original Qwen3-VL-4B-Instruct layers. Blue denotes the layers from the original model, and green denotes replaced layers, where annotations indicate the attention variant used after searching. $\ast$ indicates additional FFN pruning in the corresponding layer.}
  \label{fig:arch_viz}
\end{figure}

The MOSAIC-4B architecture achieves substantial computational savings through the strategic distribution of heterogeneous operators across the network depth. According to \cref{fig:arch_viz}, we found that it reveals several consistent structural patterns. Specifically, SWA is allocated to the shallow layers, suggesting that the early stages of the network primarily attend to local contextual information. The emergence of this pattern provides a strong validation of the MOSAIC method. By independently recovering expert-designed architectural heuristics solely through data-driven multi-objective optimization, the framework confirms the efficacy of local window mechanisms for capturing initial contextual representations \cite{xu2025mswa, beltagy2020longformer, jiang2023mistral}. The middle and later layers are predominantly assigned linear attention mechanisms, such as KDA and GDN, which account for the majority of the total latency reduction. The framework preferentially retains MLA in the tail layers, indicating that global information aggregation is essential for generating high-quality representations in the final stages. We further validate these consistent structural distribution patterns across a range of target speedup ratios, with comprehensive visualizations of the resulting architectures provided in \cref{sec:appendix_c}.

\section{Conclusion}
\label{sec:conclusion}

In this paper, we propose MOSAIC, a hardware-aware method that automates the transition from homogeneous VLMs to optimized heterogeneous architectures via adaptive inter-layer composition. MOSAIC formulates architecture selection as a multi-objective MIP problem and identifies optimal configurations within a diverse search space of efficient operators. To bridge the performance gap caused by structural transitions, we introduce a two-stage distillation method. The resulting model, MOSAIC-4B, effectively matches the performance of its Qwen3-VL-4B-Instruct teacher across 19 benchmarks while delivering a $1.76\times$ prefilling speedup and a $2.54\times$ decoding acceleration. Remarkably, these gains are achieved with less than 2\% of the original training cost. These findings highlight the efficacy of automated heterogeneous design for building efficient vision-language models.

{
    \small
    \bibliographystyle{ieeenat_fullname}
    \bibliography{main}
}

\clearpage
\setcounter{page}{1}
\appendix

\section{Details for the Search Space}
\label{sec:appendix_a}

In this section, we provide the comprehensive architectural specifications for the candidate subblocks within the search space.
For the Qwen3-VL-4B-Instruct teacher model employed in our primary experiments, the search space is meticulously designed to ensure structural compatibility with the parent model while offering a diverse range of efficiency-accuracy trade-offs.
To maintain consistency with the teacher's residual stream, the hidden dimension for all attention and FFN subblocks is fixed at $2560$.

The search space for attention modules encompasses five distinct mechanism types. Their configurations either inherit the settings from the teacher's GQA or follow the default hyperparameters from the representative \texttt{FLA}\footnote{\url{https://github.com/fla-org}} library for efficient sequence modeling:
\begin{itemize}[leftmargin=4mm]
    \item Grouped-Query Attention (GQA): This configuration utilizes the original GQA mechanism of the teacher model, featuring 32 attention heads and 8 key-value heads.
    \item Gated Delta Network (GDN): Following the architecture in \cite{yang2024gated}, this variant employs a head dimension of 128 and an expansion factor of 2 for the value states.
    \item Kimi Delta Attention (KDA): Based on the design in \cite{team2025kimi}, this mechanism uses a head dimension of 128 and an expansion factor of 1 for the value states.
    \item Multi-head Latent Attention (MLA): Adopting the configuration from \cite{deepseek2024v2}, this variant utilizes a 128-dimensional value head and 192-dimensional key/query heads. It leverages LoRA-style bottlenecks for queries with a rank of 64 and key-values with a rank of 512.
    \item Sliding Window Attention (SWA): This setup retains the teacher model's GQA parameters but restricts the attention scope to a sliding window of length 1024.
\end{itemize}

For the FFN subblocks, we employ a structured pruning approach to optimize the intermediate dimension of the SwiGLU layers.
The intermediate dimensions of all candidate subblocks are aligned to multiples of 256 or powers of two to maximize hardware execution efficiency.
Specifically, the search space includes seven scaling ratios: $\{100\%, 84\%, 74\%, 50\%, 21\%, 11\%, 0\%\}$.
These correspond to the setting $d = \{9728, 8192, 7168, 4864, 2048, 1024, 0\}$ of the intermediate dimensions, respectively, where a dimension of 0 represents the removal of the FFN subblock.

To provide a warm start for the subblocks in the search space, we initialize these alternative subblocks by leveraging the pretrained weights of the teacher model. Specifically, the GQA and SWA subblocks are initialized directly from the teacher model as they maintain structural compatibility. For the GDN and KDA linear attention variants, we initialize the head projection matrices by performing mean-pooling over the teacher's corresponding projection matrices to align the feature dimensions. To initialize the low-rank projection matrices in the MLA subblocks, we apply Singular Value Decomposition (SVD) to the teacher's projection matrices and utilize the decomposed results to populate the latent representations. Finally, for FFN subblock initialization, we reduce the teacher's intermediate dimensions by performing structured channel pruning, where channels are randomly pruned to meet the target dimensions. This strategy ensures that candidate blocks inherit foundational knowledge, significantly reducing the difficulty of the subsequent recovery stages.

\section{Additional Implementation Details}
\label{sec:appendix_b}

In this section, we provide the additional implementation details for the experiments in \cref{sec:exp}.

\subsection{Details of Blockwise Latency Evaluation}
\label{sec:appendix_b1}

\begin{figure*}[t]
    \centering
    \begin{subfigure}{0.48\textwidth}
        \centering
        \includegraphics[width=\textwidth]{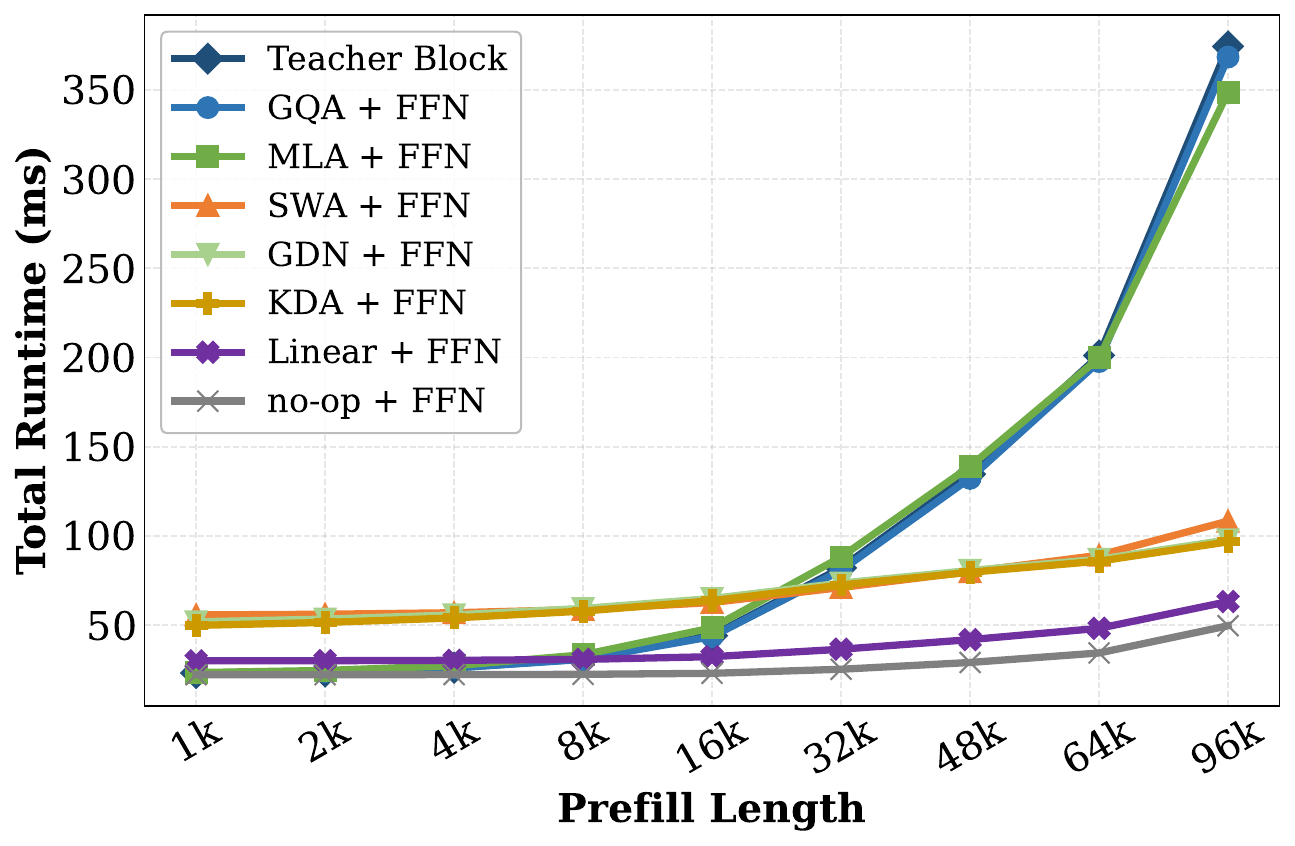}
    \end{subfigure}
    \hfill
    \begin{subfigure}{0.48\textwidth}
        \centering
        \includegraphics[width=\textwidth]{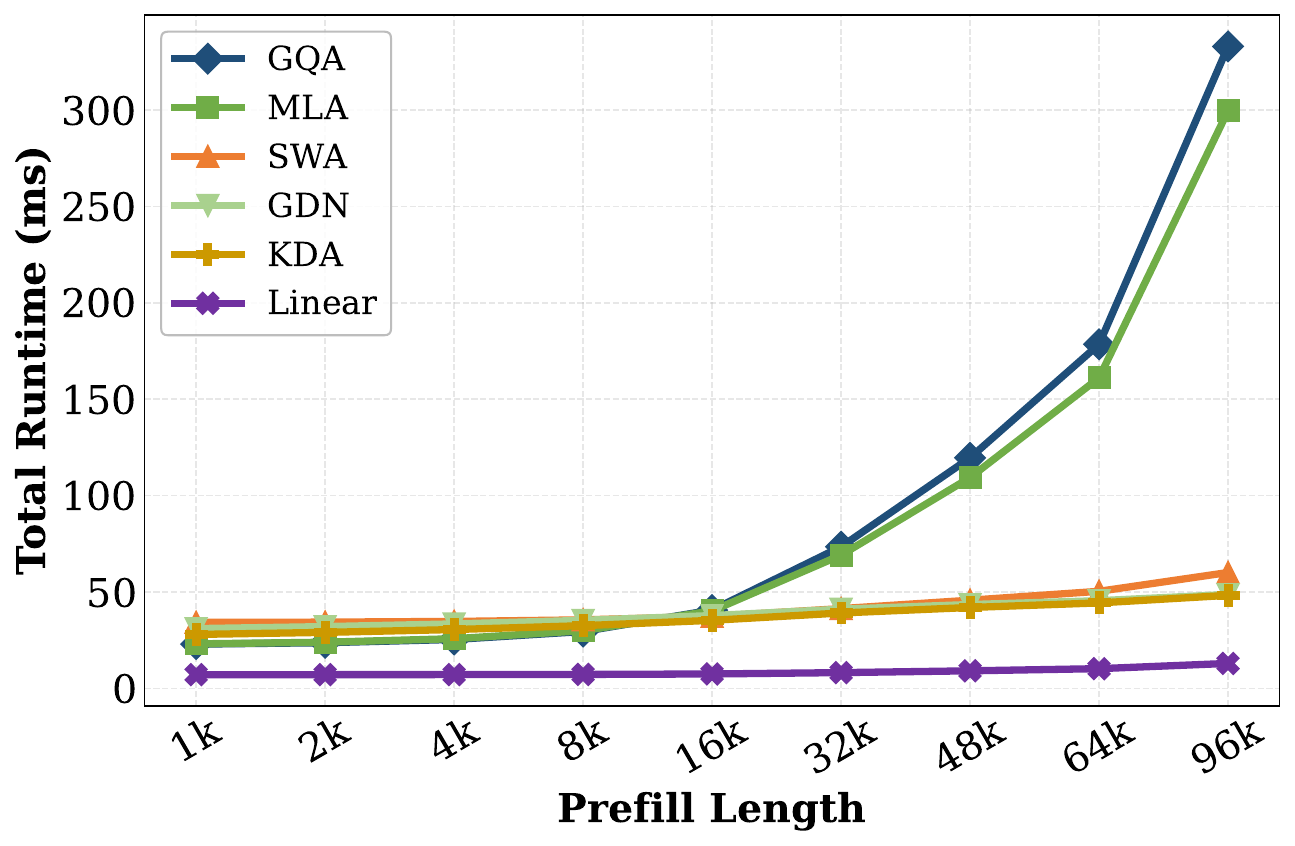}
    \end{subfigure}
    \caption{\textbf{Hardware performance analysis of various attention variants.}
    \textbf{Left:} Profiling of block-wise latency across different hardware stages, illustrating the specific time breakdown.
    \textbf{Right:} Comparison of scaling efficiency across prefill lengths from 1k to 96k (FFN cost excluded).}
    \label{fig:block_latency}
\end{figure*}

To establish the latency constraints $\mathsf{runtime}(l,i)$ for the multi-objective MIP problem, we independently profile the latency of each candidate block under realistic inference conditions. To isolate the true hardware execution time, we employ static key-value (KV) cache pre-allocation and leverage CUDA Graphs during the decoding phase, effectively eliminating dynamic memory allocation overheads and CPU dispatch bottlenecks. Specifically, we evaluate prefill lengths scaling from 1k up to 96k tokens, using a batch size of 16 and a decode length of 256 tokens. As illustrated in \cref{fig:block_latency}, across these extensive context lengths, we observe that the latency of standard GQA increases significantly, inherently bottlenecked by the computational complexity of full attention matrix multiplications. In contrast, SWA exhibits highly efficient scaling, as its computational overhead is strictly bounded by the fixed window size. Furthermore, MLA demonstrates consistent latency reductions through its low-rank KV compression, while linear attention variants, including GDN and KDA, maintain near-constant decoding latency and minimal memory footprints owing to their recurrent state mechanisms. Consequently, this rigorous profiling yields the precise, deterministic latency metrics required to populate the $\mathsf{runtime}(l,i)$ constraint, thereby accurately guiding the multi-objective MIP search process. We also provide supplementary profiling of attention-only latency in \cref{fig:block_latency}.

\subsection{Training Details}
\label{sec:appendix_b2}

Training proceeds through three stages: blockwise distillation, global off-policy
distillation, and dual-teacher on-policy distillation. The entire training
procedure is conducted on 64$\times$ NVIDIA H800-80GB GPUs over 172 hours,
amounting to approximately 11K GPU hours in total.

For blockwise distillation, we utilize 80K samples drawn from
LLaVA-OneVision-1.5-Midtrain~\cite{an2025llava}, a general-purpose VLM
captioning dataset commonly employed as multimodal pretraining data. Each block
is trained for one epoch with the MSE loss. This round of distillation serves as
an effective warm-up for the blocks, preventing inaccurate module scores that
may arise from discrepancies in parameter inheritance.

The global off-policy distillation in the parameter recovery stage proceeds in
two sub-phases.
(\romannumeral1)~\emph{Pre-training KD}: We distill from Qwen3-VL-4B-Instruct
using approximately 10M samples of pure text from ClimbMix~\cite{diao2025nemotron},
augmented with 10M image-text pairs from the LLaVA-OneVision-1.5~\cite{an2025llava}
midtrain split. For optimization, we employ AdamW with $\beta_{1}{=}0.9$, $\beta_{2}{=}0.95$, and a weight decay of $0.1$. The learning rate and effective batch size are set to $1{\times}10^{-5}$ and 512, respectively. The distillation coefficient is $\alpha{=}0.1$.
(\romannumeral2)~\emph{Instruction KD}: We continue distillation on an
instruction mixture comprising LLaVA-OneVision-1.5-Instruct, LLaVA-Video-178K,
CLEVR, NExT-QA, PerceptionTest, STAR, and
VSI-590K~\cite{yi2019clevret,xiao2021next,patraucean2023perception,zohar2024video,yang2025cambrian},
totaling approximately 10M samples.

For the dual-teacher on-policy distillation in the parameter recovery stage, we
combine MMFineVision~\cite{lin2026mmfinereason} with the data used in Stage~1, \ie, global off-policy distillation,
at a 1:1 ratio, yielding approximately 2M samples in total.
For each prompt, the SFT target is a response generated by Qwen3-VL-235B, while
the logits teacher is Qwen3-VL-4B-Instruct. We set the learning rate to
$5{\times}10^{-6}$ with $\beta=0.5$.

We select the above datasets because they have undergone high-quality cleaning
and careful mixture-ratio tuning, and have been successfully used to train
strong open-source models. The total training cost is less than 2\% of the
training cost of the original Qwen3-VL-4B-Instruct.

\subsection{Details of Benchmark Evaluation}
\label{sec:appendix_b3}

For the evaluation of image and video benchmarks, we employ \texttt{VLMEvalKit}\footnote{\url{https://github.com/open-compass/vlmevalkit}} as the evaluation framework. To ensure a fair comparison, we apply a unified set of generation hyperparameters across all models: the temperature is set to $0.7$, the maximum number of new tokens to $4{,}096$, the repetition penalty to $1.0$, the presence penalty to $1.5$, top-$p$ to $0.8$, and top-$k$ to $20$. For text-only benchmarks, we conduct evaluations using \texttt{OpenCompass}\footnote{\url{https://github.com/open-compass/opencompass}}, with the maximum input length uniformly capped at 96K tokens.

\subsection{Details of Model Latency Evaluation}
\label{sec:appendix_b4}
\begin{figure}[t]
    \centering
    \includegraphics[width=\columnwidth]{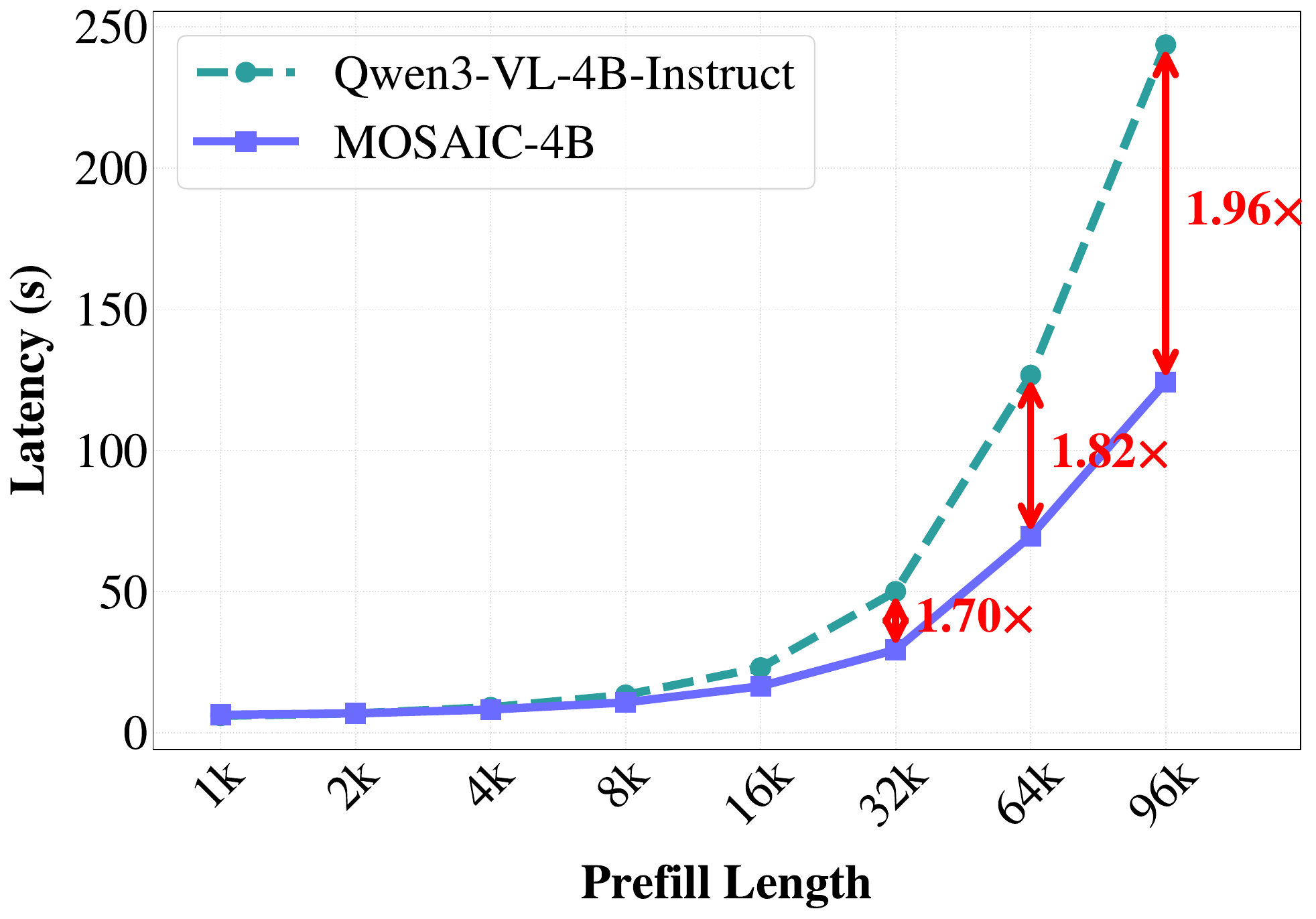}
    \caption{Comparison of end-to-end inference latency. MOSAIC-4B achieves substantially higher efficiency than the Qwen3-VL-4B-Instruct teacher model.}
    \label{fig:model_latency}
\end{figure}

To evaluate the real-world latency and throughput of the derived MOSAIC-4B model, we deploy it using vLLM~\cite{kwon2023efficient}, a highly optimized open-source inference engine. vLLM achieves high throughput by utilizing custom PagedAttention kernels to efficiently manage KV caching and mitigate memory fragmentation. To seamlessly deploy the specific heterogeneous, per-layer paradigms generated by our MOSAIC search space, we further extend vLLM's backend.
Specifically, we develop a unified decoder layer that dynamically dispatches to specialized operators based on the searched configuration. Built upon PagedAttention, the engine natively supports GQA with variable KV-head ratios and SWA. We extend this paging capability to MLA to accommodate its low-rank KV compression scheme. To support GDN and KDA linear attention, we incorporate their internal recurrent states into a hybrid state management system. This allows the memory scheduler to dynamically allocate and page these states identically to standard KV caches. Consequently, by replacing isolated implementations with native, hardware-aware operators, the extended engine fully supports the latency evaluation and practical deployment of any heterogeneous architecture derived from our MIP formulation. \cref{fig:model_latency} illustrates the end-to-end inference latency of MOSAIC-4B and Qwen3-VL-4B-Instruct across different prefill lengths, where MOSAIC-4B achieves up to a $1.96\times$ latency speedup at 96k tokens.

\section{Additional Experiments}
\label{sec:appendix_c}
In this section, we provide additional experimental results.

\subsection{Comparison with Heterogeneous VLMs}
\label{sec:appendix_c1}

To further demonstrate the effectiveness of MOSAIC, in addition to the results in \cref{tab:main_image}, we compare MOSAIC-4B against three representative heterogeneous VLMs: MatVLM~\cite{li2025matvlm}, InfiniteVL~\cite{tao2025infinitevl}, and Cobra~\cite{zhao2025cobra}. All models are evaluated on the same benchmark suite described in \cref{sec:setup}, with results aggregated into three modality categories: image understanding (Img), video understanding (Vid), and text understanding (Txt). As shown in \cref{tab:hetero_compare}, MOSAIC-4B achieves the best performance across all three modalities by a substantial margin. Notably, Cobra and MatVLM lag significantly behind, suggesting that naive manual hybridization strategies---whether through simple module concatenation or rule-based layer substitution---fail to capture the nuanced, layer-wise architectural preferences that automated search can exploit. These results confirm that principled heterogeneous architecture search yields considerably superior designs compared to the hand-designed architecture.

\begin{table}[t]
  \caption{Performance and decoding efficiency comparison between MOSAIC variants and representative heterogeneous VLMs.
  Img, Vid, and Txt denote the average performance (\%) on image, video, and text understanding benchmarks, respectively.
  TPOT denotes the decoding acceleration (time per output token) relative to the corresponding homogeneous baseline.
  \textbf{Bold}: best; ${}^\dagger$: second best among competing models.}
  \label{tab:hetero_compare}
  \centering
  \setlength{\tabcolsep}{5pt}
  \resizebox{\columnwidth}{!}{%
  \begin{tabular}{@{}lc|cccc@{}}
    \toprule
    Model & TPOT & Img & Vid & Txt & Overall \\
    \midrule
    \rowcolor{gray!15}
    Qwen3-VL-4B-Instruct & $1.00\times$ & 78.5 & 57.2 & 83.1 & 72.9 \\
    \midrule
    Cobra~\cite{zhao2025cobra}         & $\mathbf{4.83\times}$ & 56.2 & 35.8 & 60.4 & 50.8 \\
    MatVLM~\cite{li2025matvlm}         & $1.21\times$ & 62.1 & 41.3 & 64.5 & 56.0 \\
    InfiniteVL~\cite{tao2025infinitevl} & $4.21\times$ & 69.8 & 48.3 & 72.1 & 63.4 \\
    \midrule
    \textbf{MOSAIC-4B}       & $2.54\times$ & \textbf{77.9} & \textbf{56.4} & \textbf{80.8} & \textbf{71.7} \\
    \textbf{MOSAIC-4B-Flash}   & $4.42^\dagger\times$ & $75.1^\dagger$ & $53.8^\dagger$ & $77.3^\dagger$ & $68.7^\dagger$ \\
    \bottomrule
  \end{tabular}}
\end{table}

\subsection{Performance of Faster Configurations}
\label{sec:appendix_additional}
\begin{figure*}[t]
    \centering
    \begin{subfigure}{0.48\textwidth}
        \centering
        \includegraphics[width=\textwidth]{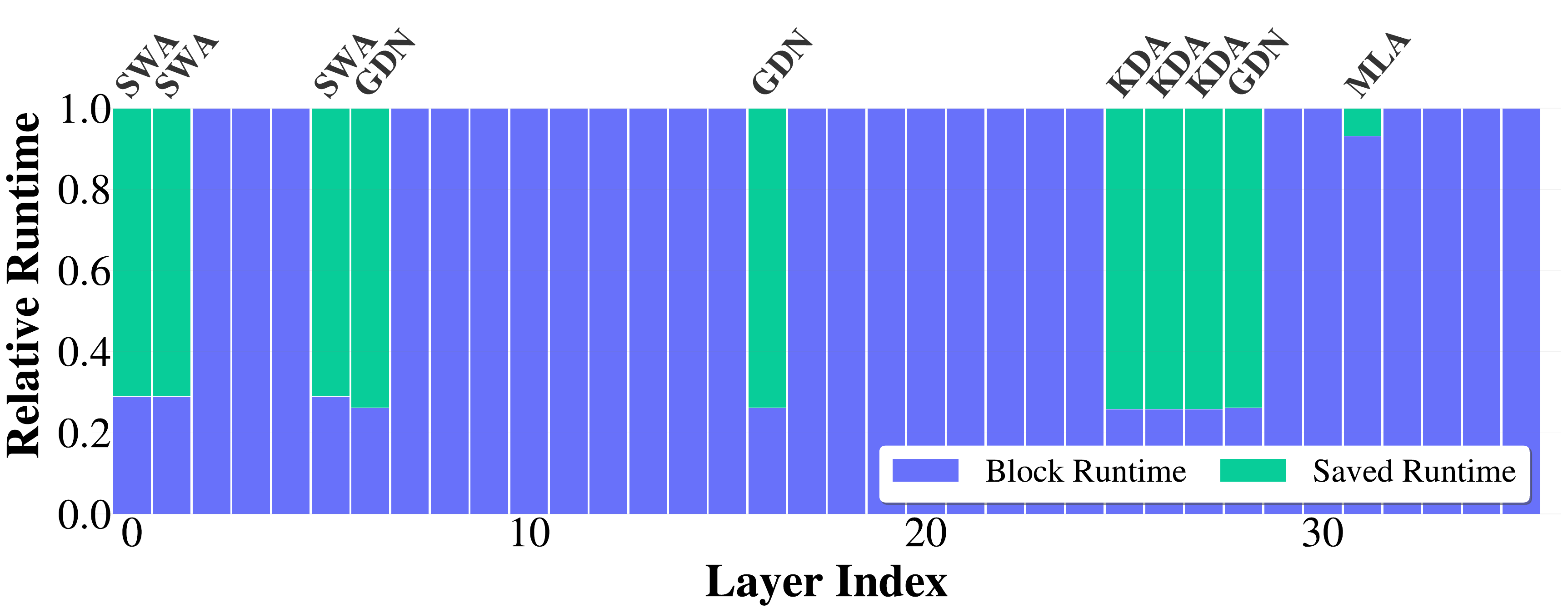}
        \caption{$1.2\times$ speedup.}
    \end{subfigure}
    \hfill
    \begin{subfigure}{0.48\textwidth}
        \centering
        \includegraphics[width=\textwidth]{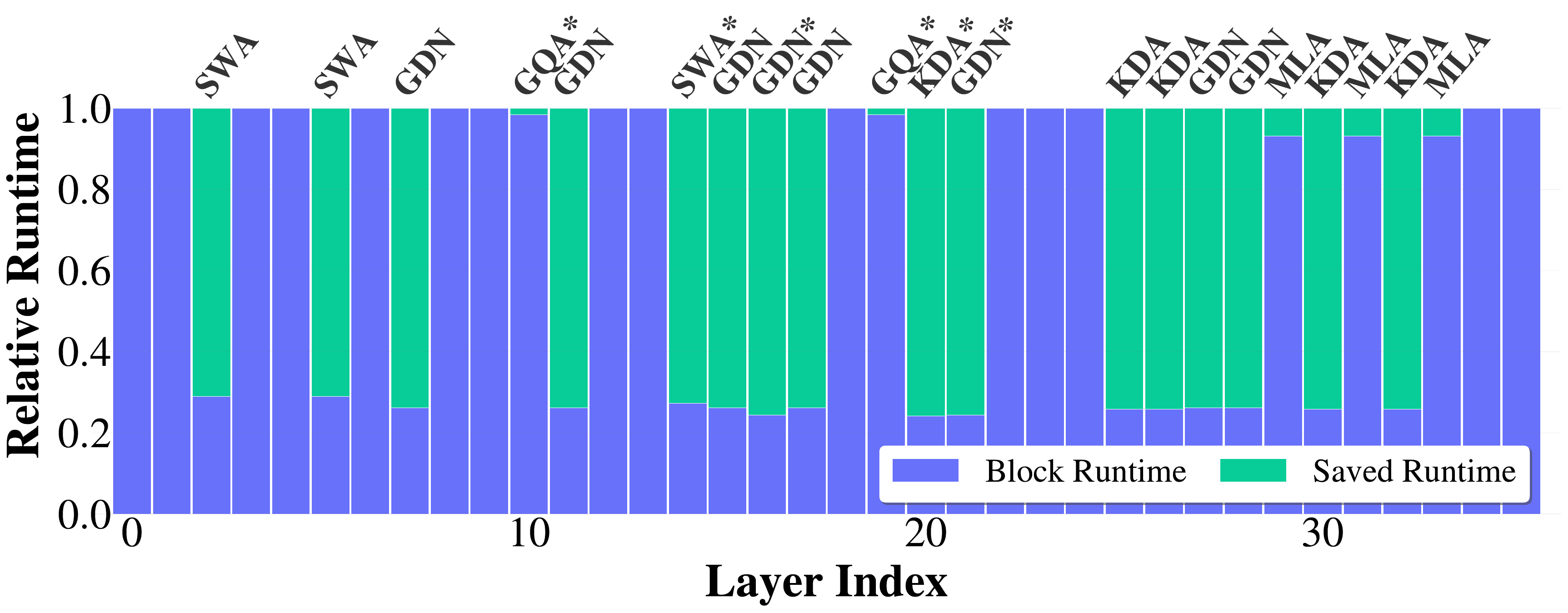}
        \caption{$1.5\times$ speedup.}
    \end{subfigure}

    \vspace{1em}

    \begin{subfigure}{0.48\textwidth}
        \centering
        \includegraphics[width=\textwidth]{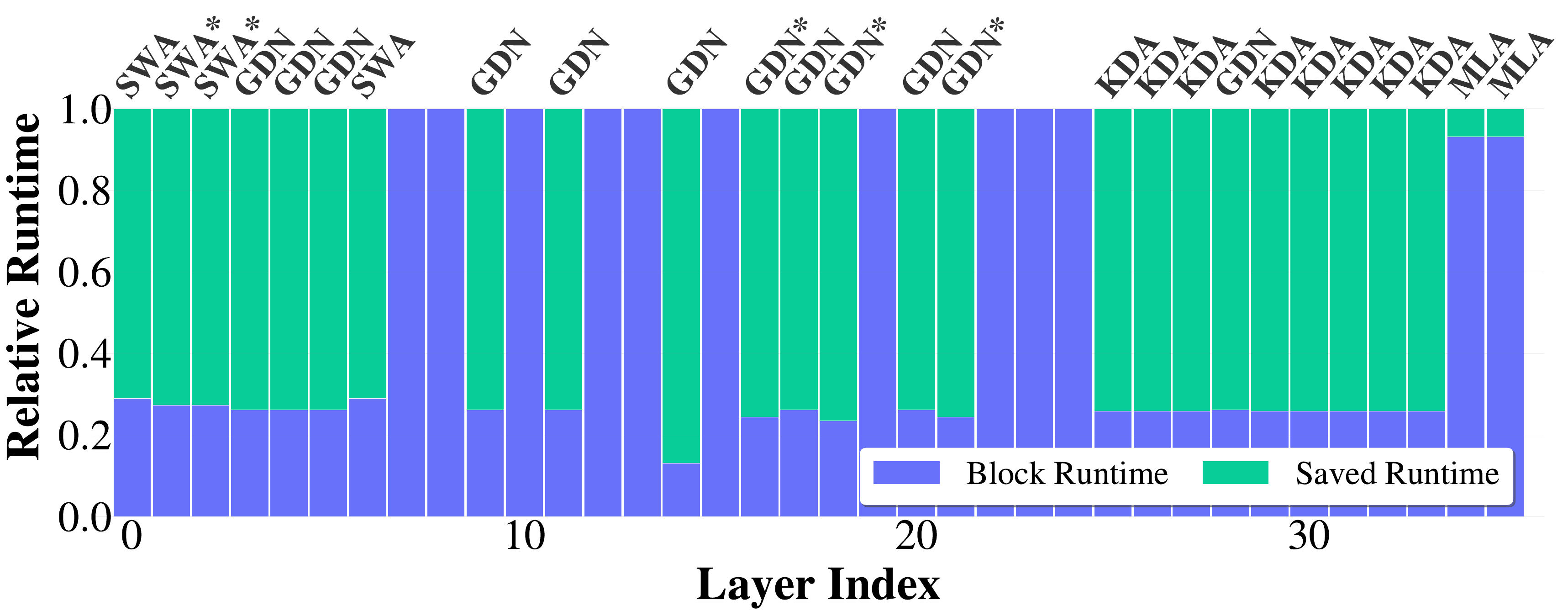}
        \caption{$2\times$ speedup.}
    \end{subfigure}
    \hfill
    \begin{subfigure}{0.48\textwidth}
        \centering
        \includegraphics[width=\textwidth]{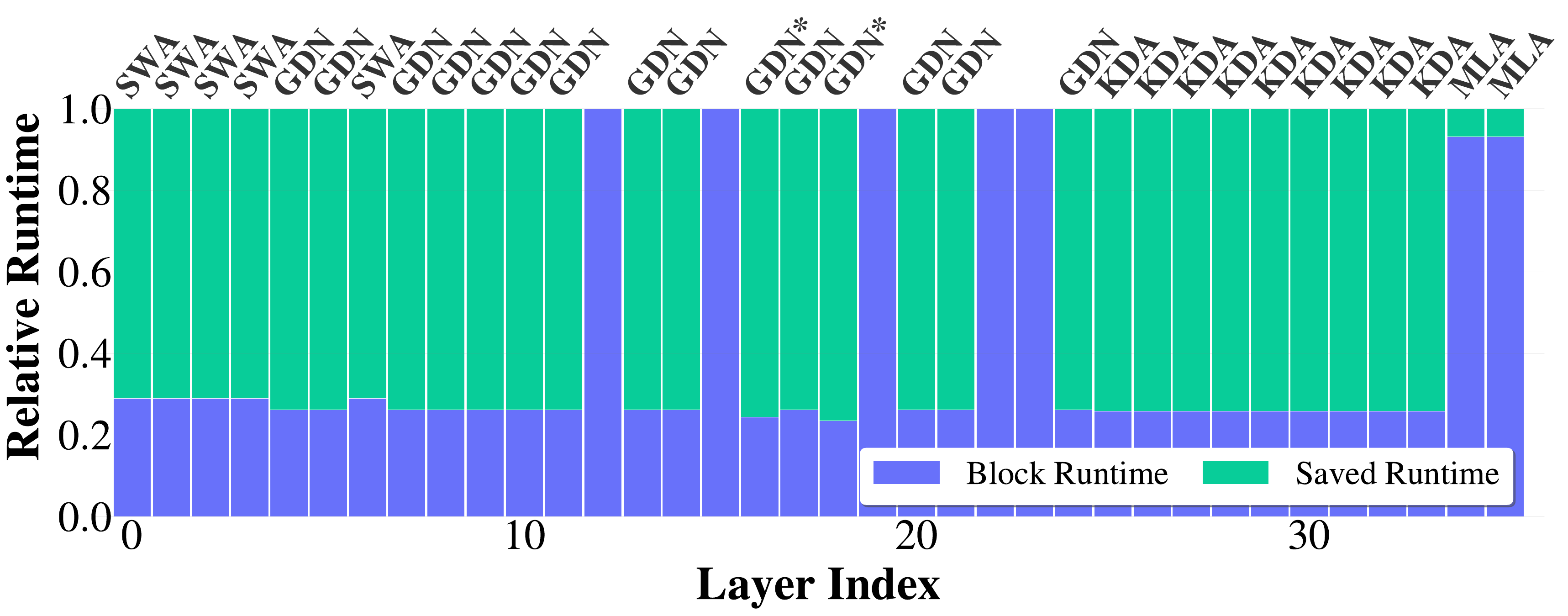}
        \caption{$2.5\times$ speedup.}
    \end{subfigure}

    \caption{The runtime reduction of layer-wise architectural replacements in MOSAIC under varying latency constraint settings ($1.2\times$ to $2.5\times$ speedup), relative to the original Qwen3-VL-4B-Instruct layers. Blue denotes the layers from the original model, and green denotes replaced layers, where annotations indicate the attention variant used after searching. $\ast$ indicates additional FFN pruning in the corresponding layer.}
    \label{fig:four_search_results}
\end{figure*}

To further investigate the flexibility of the MOSAIC framework, we evaluate a speed-oriented variant, MOSAIC-4B-Flash, derived from the architectural configuration discovered under a $2.0\times$ speedup constraint, which achieves competitive inference speed with existing heterogeneous VLMs. This model has undergone the same two-stage recovery process as the MOSAIC-4B. The corresponding model architecture is shown in \cref{fig:four_search_results} (c). As reported in \cref{tab:hetero_compare}, MOSAIC-4B-Flash achieves a measured decoding acceleration of $4.42\times$, substantially surpassing the $2.54\times$ speedup of the standard MOSAIC-4B.

Experimental results indicate that this increased efficiency comes at a relatively modest cost to model quality. Specifically, MOSAIC-4B-Flash exhibits a performance decrease of approximately 3.0 points in the overall average score compared to the standard model, while still maintaining a clear performance advantage over other heterogeneous VLMs at comparable speed levels. These findings suggest that the MOSAIC framework can effectively navigate the Pareto frontier between inference efficiency and task accuracy, potentially providing a range of optimized candidates to meet diverse hardware deployment requirements.

\subsection{Experiments on Architecture Patterns}
In \cref{sec:discover}, we analyze the architectural patterns of the MOSAIC-4B model searched under a global latency budget constrained to a $1.5\times$ speedup relative to the Qwen3-VL-4B-Instruct baseline. In this section, we evaluate the analysis results of discovered architectural patterns by visualizing the configurations identified by MOSAIC across a wide spectrum of latency constraints, ranging from $1.2\times$ to $2.5\times$ speedup relative to the Qwen3-VL-4B-Instruct baseline.

As illustrated in \cref{fig:four_search_results}, several fundamental structural patterns remain stable across these different latency constraints. Specifically, SWA is consistently allocated to the shallow layers of the network, reinforcing the conclusion that early stages primarily benefit from local contextual information for initial representation. This finding aligns with prior works \cite{xu2025mswa, beltagy2020longformer, jiang2023mistral} supporting the efficacy of local window mechanisms in initial layers for capturing fundamental contextual features. Meanwhile, linear attention variants like KDA and GDN predominantly occupy the middle layers, serving as the primary driver for latency reduction across all evaluated constraints. In the deep layers, the framework preferentially retains MLA or standard GQA, suggesting that global information aggregation remains essential for the final stages of the model.

These findings align with the structural patterns observed in \cref{sec:discover}. Since these architectural patterns were recovered through data-driven multi-objective optimization rather than manual intervention, this indicates that such heterogeneous distributions may offer a favorable balance for vision-language modeling. Such discovered architecture patterns may offer useful architectural intuitions and serve as a potential reference for future explorations in efficient multimodal design.

\end{document}